\documentclass[runningheads]{llncs}

\usepackage{eccv}

\usepackage{eccvabbrv}

\usepackage{graphicx}
\usepackage{booktabs}
\usepackage{multirow}
\usepackage{xcolor}
\usepackage{colortbl}
\usepackage{nicematrix}
\usepackage[accsupp]{axessibility}  
\usepackage{pifont}

\usepackage{pdfpages}

\usepackage{hyperref}

\usepackage{orcidlink}

\begin{document}

\title{Image-guided topic modeling for interpretable privacy classification}


\author{Alina Elena Baia\inst{1}\orcidlink{0000-0001-5553-776X} \and
Andrea Cavallaro\inst{1,2}\orcidlink{0000-0001-5086-7858} 
}

\authorrunning{A.E.~Baia and A.~Cavallaro}

\institute{Idiap Research Institute, Switzerland, \and
École Polytechnique Fédérale de Lausanne, Switzerland\\
\email{\{alina.baia, a.cavallaro\}@idiap.ch}
}

\maketitle

\begin{abstract}Predicting and explaining the private information contained in an image 
in human-understandable terms is a complex and contextual task. This task is challenging even for large language models. To facilitate the understanding of privacy decisions, we propose to predict image privacy based on a set of natural language content descriptors. These content descriptors are associated with privacy scores that reflect how people perceive image content. We generate descriptors with our novel Image-guided Topic Modeling (ITM) approach. ITM leverages, via multimodality alignment, both vision information and image textual descriptions from a vision language model.
We use the ITM-generated descriptors to learn a privacy predictor, Priv$\times$ITM, whose decisions are interpretable by design. Our Priv$\times$ITM classifier outperforms the reference interpretable method by 5 percentage points in accuracy and performs comparably to the current non-interpretable state-of-the-art model.

\end{abstract}

\section{Introduction}
\label{sec:intro}

Images shared online may reveal personal information, such as location, social habits, and sexual, political and religious orientations~\cite{GIP}. This information can be aggregated and (mis)used without the person's informed consent. 
Warning users about potentially sensitive content prior to sharing their images would help avoid unwanted privacy violations. However, training an image-privacy classifier that highlights why a prediction was made is challenging as privacy is a subjective and context-dependent concept. The individuals' views on privacy are influenced by various factors, such as cultural background and life experiences~\cite{Cultural_differences, Li2022, privacy_experiences}.

Identifying private information in images is tackled as a privacy 
prediction task~\cite{Zerr_traditional,Buschek_metadata,Tonge_cnn,PrivacyAlert, Dynamic_fusion,GIP,stoidis2022content} or as a recommendation of personalized settings~\cite{tag-to-protect,iPrivacy,YU_leveraging, Personalized_privacy,orekondy_68_attributes,dammu2021explainable}. Privacy classification models may be trained with hand-crafted visual features~\cite{Zerr_traditional},  a combination of visual features and metadata~\cite{Buschek_metadata}, deep visual features~\cite{Tonge_cnn}, 
fusion of deep visual features and tags~\cite{PrivacyAlert} or objects information, scene context and tags~\cite{Dynamic_fusion}. 
Works also explored personalized privacy classification using image tags~\cite{tag-to-protect}, user feedback and privacy preferences~\cite{Personalized_privacy,orekondy_68_attributes},
privacy patterns of groups of similar users in social media sites~\cite{zhong2017group}, or the combination of image content sensitiveness and user trustworthiness~\cite{YU_leveraging}. 
However, the above methods do not explain the specific privacy-related elements, thus limiting a user's ability to make informed decisions about the risks of image sharing. While post-hoc explanation methods may be used to generate relevance maps 
that highlight image regions that are important for a decision~\cite{LIME, selvaraju2017grad, simonyan2013saliency}, no information is given on how and why those pixels influence the prediction. 

We aim to make the decision-making process understandable through natural language.
Concepts bottleneck models (CBMs)~\cite{koh2020concept,losch2019interpretability, zhou2018interpretable, yun2022vision} use a linear combination of interpretable concepts to make predictions. 
CBMs can be constructed without human annotations by eliciting domain knowledge from LLMs~\cite{yang2023labo, oikarinen2023labelfree,yan2023robustmedical} or knowledge bases~\cite{yuksekgonul2022post}. LLMs are prompted to describe a {category} (e.g.~shape, color, patterns) or to list important features to build a set of concepts (i.e.~concise descriptors). While LLMs perform well on standard computer vision tasks, they are still inadequate in comprehensively listing abstract image attributes, such as those making an image private\footnote{See prompting examples and privacy attributes in Appendix A and K, respectively.}. Human intervention is needed to tackle this issue, for example, via manual refinement of attributes or guided prompts which is time-consuming and limits the scalability and automation of the process. 

To address these limitations, we propose Image-guided Topic Modeling\footnote{Topic Modeling is a technique to discover latent topics (groups of frequently co-occurring words representing themes or ideas) in a large corpus of text data~\cite{LDA, NMF,grootendorst2022bertopic}.} (ITM), a new approach for interpretable image classification of complex and abstract tasks that does not rely on human-specified image attributes. ITM produces human-understandable content descriptors, which can be used to make predictions as well as to explain them, using a Large Vision Language Model (LVLM). We improve topic representation by discovering topics from deep tags extracted from image textual descriptions within clusters of similar images. Next, we merge the topics’ word representations obtained within a cluster into a content descriptor via visual information of the cluster.  We use the set of descriptors that summarize the content in a dataset to train a linear classifier on the image-descriptor association scores computed with a pretrained multimodal alignment model. The image-descriptor association scores indicate how strongly a descriptor is associated with an image, providing a quantitative measure of their semantic alignment. The learned weight matrix of the classifier reflects the relevance of each content type in the final classification and can be used to interpret the model's decisions. We show that ITM\footnote{Code is available at \href{https://github.com/idiap/itm}{https://github.com/idiap/itm}} enables the construction of interpretable-by-design classifiers that outperform existing interpretable methods and obtain comparable results with non-interpretable models.
Because a direct comparison with previous methods is not feasible due to the fundamental differences in the methods' design, we also propose a new (non-interpretable) baseline  SVM$\times$IB, a support vector machine trained on image embeddings extracted from a multimodal  model~\cite{girdhar2023imagebind}.   
SVM$\times$IB  outperforms the current state-of-the-art model and sets a new benchmark for the privacy classification problem.

\section{Related work}
\label{sec:related_work}

\noindent \textbf{Black-box methods}. 
Methods for image privacy prediction use objects and convolutional features~\cite{pcnh}, a fine-tuned transformer-based model (BERT) with user-defined and automatically generated image tags~\cite{PrivacyAlert}, or image and tags fusion with two-stream transformers (ViLBERT)~\cite{PrivacyAlert}. A knowledge graph that encodes the relationship between objects and privacy labels can also be used~\cite{GIP}. A dynamic region-aware graph network adaptively models the correlation between relevant image regions with a self-attention mechanism and no pretrained object detectors~\cite{drag}.
Scene information can be fused with object co-occurrence and cardinality to train a graph-based  classifier~\cite{stoidis2022content}. Object detection, scene and tags-based classifiers can be fused through the weights of class probability distributions based on the per-image reliability of fine-tuned unimodal classifiers achieving state-of-the-art results~\cite{gate_multi_modal_fusion}.

\noindent \textbf{Explanations}. 
Methods that generate human-interpretable explanations use regular expressions to describe privacy decisions with natural language. These decisions are based on the late fusion of object and people detection, location and scene information, and explicit adult content~\cite{dammu2021explainable}. 
This framework (similar  to~\cite{saliency_privacy_map, Jiao2020IEyePI, gate_multi_modal_fusion, Darya_8ps}) relies on prior knowledge using scene recognition, face and nudity detection, informed by studies on privacy perception~\cite{li_28_attributes,orekondy_68_attributes} and privacy classification~\cite{Tonge_scene_context_2018}.
PEAK~\cite{peak} explains privacy predictions using latent topics identified from image tags using Topic Modeling (TM) via non-negative matrix factorization, which decomposes the image-tag matrix into image-topic and topic-tag matrices.
The term weighting method  (TF-IDF) is used to measure the presence of tags in images and to ultimately compute the image-topic association scores to obtain the topic vectors. These vector representations are used to train a RandomForest classifier.  
PEAK is interpretable, despite not being originally presented as such by the authors, who proposed using post-hoc explanation methods to explain its decisions (i.e. SHAP~\cite{shap} tree explainer).
The most relevant topics for the prediction 
are used to form decision explanations with a predefined sentence structure. However, incorrect tags might be assigned to images either by automatic tagging systems (i.e.~hallucinations) or by humans. This leads to imprecise TF-IDF scores and topic misrepresentation. 

\noindent \textbf{Privacy taxonomies, features and saliency maps}. 
Privacy taxonomies have been proposed based on user studies~\cite{li_taxonomies,li_28_attributes,orekondy_68_attributes}. A multi-task learning model can be used to identify a set of privacy-sensitive objects for privacy settings recommendations~\cite{iPrivacy}. 
Human-defined features, such as the number and probability of the presence of people, the probability of the
scene being outdoors,  the likelihood to contain sexual, medical, or violent content, can also be used independently or in addition to deep features~\cite{Darya_8ps}.
Saliency privacy maps are generated using deep and traditional features by computing pixel-level privacy scores based on the maximum private probability of any patch to which the pixel belongs~\cite{saliency_privacy_map}. 
Similarly,  a series of predefined categories of visual features are employed to detect private areas in images
and to provide interpretable privacy decisions~\cite{Jiao2020IEyePI}. 

\noindent \textbf{Novelty}.  Our method detects relevant features (descriptors) based on image content, providing a more general and flexible approach to privacy classification. Unlike~\cite{dammu2021explainable,saliency_privacy_map, Jiao2020IEyePI, gate_multi_modal_fusion, Darya_8ps,stoidis2022content}, we do not need to define prior knowledge or privacy-tailored modules.  Furthermore, unlike PEAK~\cite{peak}, and inspired by recent works on CBMs~\cite{yang2023labo, yan2023robustmedical, oikarinen2023labelfree},  we determine the image-descriptor association scores with a multimodal alignment model that maps image and text into a joint embedding space that preserves the semantic meaning between the two modalities: highly related descriptors to the image content will be close in the embedding space, thus resulting in a high association score, while unrelated descriptors will produce low alignment scores. 
Moreover, unlike PEAK~\cite{peak}, which applies TM directly to the entire set of tags, we apply TM within sets of similar images and we guide the descriptor generation by the image modality which provides richer information than the text modality, and generate a better content representation, as discussed in \cref{sec:validation}. 

\section{Interpretability by design}
\label{sec:methodology}
\begin{figure}[t]
\centering
\includegraphics[width = 1.0\textwidth]{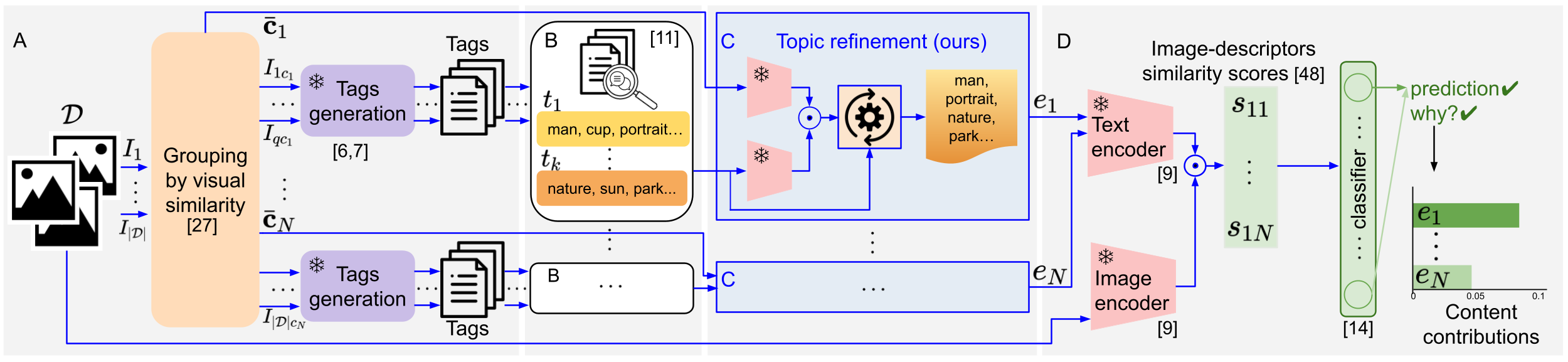}

\caption{An overview of our novel multimodal framework that enables the learning of a classifier whose decisions can be interpreted using natural language. From image tags generated within clusters, $c_j$, of visually similar images (A) topics are discovered (B) and then refined via modality alignment with the clusters' visual representation, $\bar{\mathbf{c}}_j$, to generate content descriptors, $e_j$ (C).
These descriptors are a text summary of content in a cluster, which are used as features of the images to predict image privacy and interpret the decisions (D).
Our approach differs from PEAK~\cite{peak}, which discovers topics (B) from the full tags set without image-based guidance, and GATED~\cite{gate_multi_modal_fusion}, which fuses unimodal image/text classifier outputs.
}
\label{fig:topic_modeling_inside_clusters}
\end{figure}

We propose to generate a set of content descriptors that serve as a basis for both accurate decision-making and interpretability via image-guided topic modeling (ITM). An overview of our method is shown in \cref{fig:topic_modeling_inside_clusters}. By performing topic modeling on the tags-based representation in a cluster of visually similar images, ITM identifies a set $\mathcal{E} =\{e_1, e_2, \dots, e_N$\} of $N$ multi-word content descriptors, $e_j$, from clusters, $c_j$. For each cluster $c_j$ we select multiple words from the representation of the discovered topics to create $e_j$. We then form the interpretable classifier with a fully connected layer where each of the $N$ input neurons corresponds to one $e_j$.
%

\noindent{\textbf{Content categorization.}}
We leverage topic modeling to generate content descriptors, $e_j$. However, privacy-relevant terms may be overpowered by common terms during topic discovery leading to content descriptors that lack specificity. In fact, topic modeling often struggles to distinguish similar pieces of text with different meanings\footnote{For example, the phrases \textit{picture, naked, person} and \textit{picture, person} have a high cosine similarity of 0.66 when using SentenceBERT~\cite{sentenceTransformer} embeddings.}. Furthermore, topic modeling with inaccurate text can lead to the discovery of incorrect topics.  Thus, we propose to guide the topic discovery and descriptors generation process with visual information.
To this end, we use embeddings for image representation that lie on a joint space generated with multiple modalities~\cite{girdhar2023imagebind}\footnote{
We choose ImageBind because it achieves the highest private recall in zero-shot image privacy classification. Details are available in Appendix D.}(e.g. images, text, audio). 
Multimodality training enhances a model's ability to generalize, leading to improved performance when dealing with new, unseen data such as privacy-related content that is not covered by the commonly used pretraining datasets. 
Based on these image embeddings, we group semantically similar images (that depict similar objects, scenes, actions). The joint space enables the matching of images with text, allowing us to refine content descriptors by removing words unrelated to the clusters' content.
We use density-based clustering (HDBSCAN~\cite{hdbscan}) to categorize content without the need to explicitly define the number of clusters/categories. 
This ensures that the number of clusters and their boundaries are determined by the structure of the data and promotes natural grouping, rather than specifying the number of clusters a priori. We evaluate and discuss the results of clustering in \cref{sec:validation}.

\noindent{\textbf{Image tags.}}
We proceed with image tags generation and topic discovery within each cluster to create the corresponding natural language descriptors $e_j$. 
To achieve this, we use LVLM-generated image descriptions~\cite{instructblip} to obtain image tags. With the image descriptions, we aim to capture task-relevant elements in the images people focus on.
Descriptions provide helpful information to identify a private image, such as the surroundings of an object or subject in the image (image context),  object attributes, and image atmosphere (e.g.~the overall mood is sensual and alluring). We analyzed the descriptions generated for PrivacyAlert~\cite{PrivacyAlert}: on average, descriptions have 5.50 $\pm$ 1.11 sentences and 102.01 $\pm$ 20.07  words.
As encoding long text may lead to loss of information thus reducing the performance in semantic similarity tasks, we produce a more compact textual representation by extracting keywords (the most representative terms in the text) from the descriptions to summarize the main elements of the text~\cite{vicuna2023}.  
As LVLMs are prone to hallucinations (i.e.~the model generates factually incorrect text about the input image), to improve the reliability of the generated text we use phrase grounding (i.e.~the task of identifying the object or region in the image that corresponds to a textual phrase~\cite{GLIP}). Specifically, we use an open-set object detector~\cite{liu2023groundingdino}, which detects arbitrary objects with attributes specified by natural language inputs, and we only keep keywords that are successfully grounded to their corresponding image (\textit{image tags})\footnote{Examples of image descriptions, keywords and image tags are shown in Appendix B, whereas the prompt templates for description generation~\cite{instructblip} and keywords extraction~\cite{vicuna2023} are shown in Appendix C.}.
%
%
%
%
%
%

\noindent\textbf{Image-guided topic modeling for descriptors generation.}
Next, we discover topics from the tags-based representation of images within each cluster $c_j$, and use the topics' representations to produce descriptors $e_j$ of the clusters' content.  We use BERTopic~\cite{grootendorst2022bertopic} which finds topics by clustering semantically similar documents (tags-based representation of images in our case).
This topic model generates a word representation for each topic (i.e. text-based clusters) using a variant of Term Frequency-Inverse Document Frequency (TF-IDF)~\cite{Joachims1997APA} that computes an importance score $h$ for words within a topic $t$ as:
\begin{equation}
    h_{w,t} = \|{f_{w,t}}\|\cdot \log\bigg(1 + \frac{a}{f_w}\bigg),
\end{equation}
where $f_{w,t}$ is the count of a word $w$ in a topic $t$, $f_w$ is the count of the word $w$ across all topics and $a$ is the average number of words per topic.
The $f_{w,t}$ is $L_1$-normalized to account for topic size variations.
Hence, $h_{w,t}$ models the importance of words in topics instead of individual documents.
For each topic, we select the tags with the top-10\footnote{The value of 10 was chosen based on the average number of image tags (9.69 $\pm$ 3.63) in the PrivacyAlert~\cite{PrivacyAlert} dataset.} $h_{w,t}$ scores as topic representation. We consider the tags of all topics' representations as candidates for the cluster content descriptor. As some tags might appear in the topic representation because of hallucinations (e.g. objects like \textit{chair} have been found to be frequently hallucinated~\cite{POPE_hallucinations}), we 
want the final cluster content descriptor to be relevant to the content of images in the cluster. To do this, we leverage modalities alignment in a joint embedding space: 
we remove the tags without a strong semantic alignment with the images in a particular cluster, meaning they do not accurately describe or relate to the visual content of the images.
Let $c_j$ be represented by $T$ topics $t_k$, where each topic ${t_k}= [w_{1k}, w_{2k}, \ \dots \ , w_{10k}]$ of tags, with $1 \leq k \leq T$; and let $\bar{\mathbf{c}}_j \in \mathbb{R}^d$ be the embedding representation of the centroid of cluster $c_j$.
For each tag $w_{jk}$, with  $1 \leq j \leq 10$,  $1 \leq k \leq T$, we compute an alignment 
score $r_{jk}$ as $r_{jk} = cos(\bar{\mathbf{c}}_j, \mathcal{M}(w_{jk}))$ where $\mathcal{M}$ is a multimodal alignment model (e.g. ImageBind~\cite{girdhar2023imagebind}) that maps images and text into a joint embedding space, and $\cos(\cdot)$ is the cosine similarity.
Since the same tag may appear in different topics, we remove duplicates. Note that we do not apply word singularization as, in some scenarios, this would cause a loss of meaning. For example, words like \textit{crowd} or \textit{group} will become \textit{person} or \textit{individual}. A previous study~\cite{stoidis2022content} analyzed the importance of cardinality in the \textit{person} category and observed that an image is more likely to be public if the cardinality of \textit{person} is high. 
We select 10 $w_{qp}$ to form the final content descriptor $e_j$, such that their $r_{qp}$ is in top-10 among all $r_{jk}$, with   $1 \leq j \leq 10$, $1 \leq k \leq T$.
\begin{figure}[t]
\centering
\includegraphics[width = 0.99\linewidth]{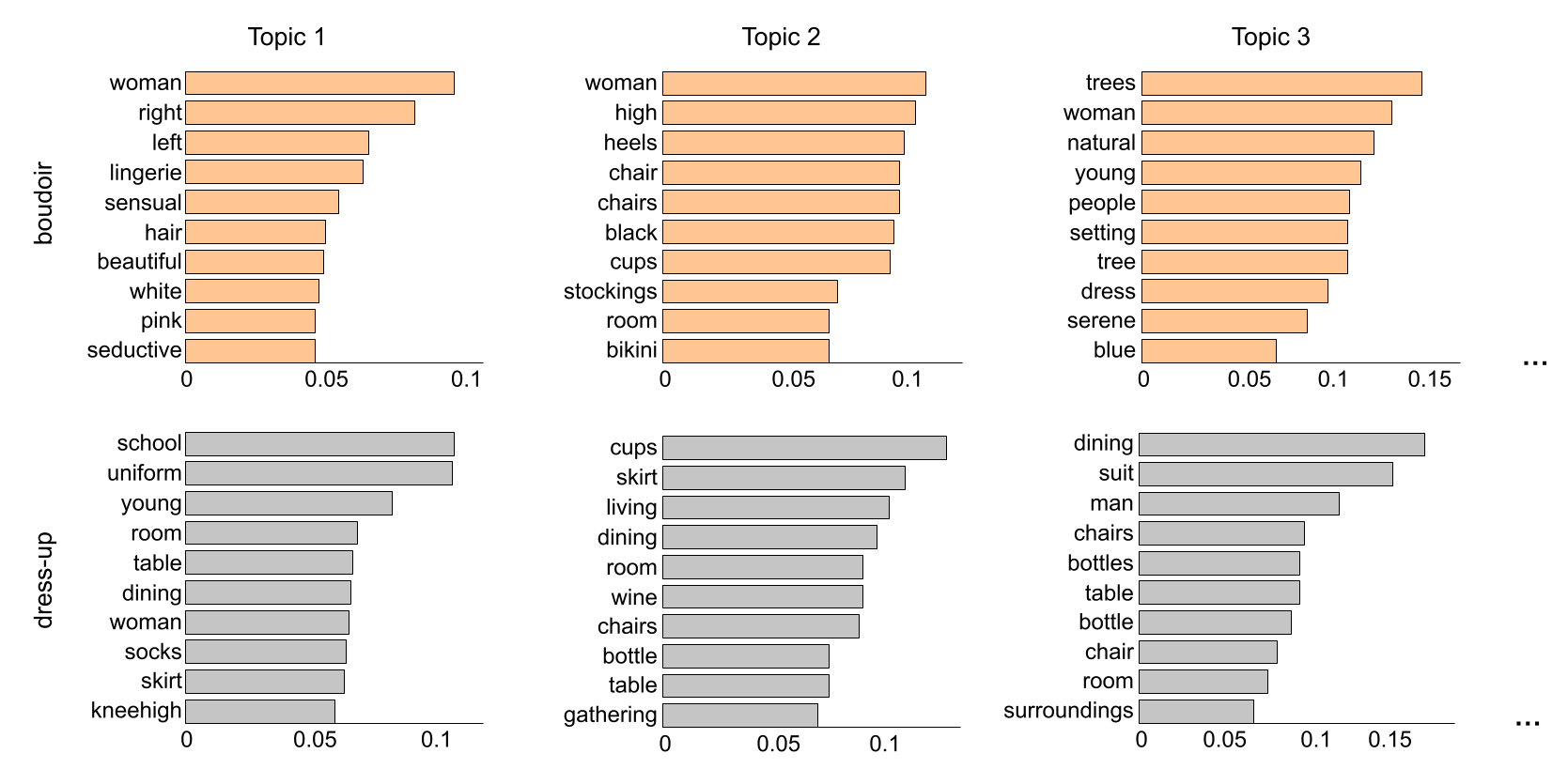}
\caption{Topics representation discovered within the image clusters \textit{boudoir} and \textit{dress-up} of PrivacyAlert~\cite{PrivacyAlert}. The x-axis shows the $h_{w,t}$ scores, and the y-axis shows the top-10 most important words for the topics according to the $h_{w,t}$ scores. These words are candidates for forming the descriptor. After applying the content-based filtering, we obtain the following content descriptor of \textit{boudoir} cluster: \textit{beautiful, seductive, lingerie, sensual, stockings, woman, black$^*$, side, dark, heels}. The descriptor of cluster \textit{dress-up} is: \textit{uniform, school, kneehigh, skirt, socks, suit, man, photo, dining, living}. Note$^*$: \textit{black} as a color of clothing. }
\label{fig:topics_for_cluster_budoir}
\end{figure}

\cref{fig:topics_for_cluster_budoir} shows the effect of the cluster-based filtering on the cluster's descriptor. For ease of identification, we name the image clusters based on their descriptors. We observe that the filtering removes objects that are often hallucinated from the representation, such as \textit{cups} and \textit{chairs}: 12 images in the cluster \textit{boudoir} out of 69 have the tag \textit{cup} and they are all hallucinations. 
Although grounding can remove hallucinated objects in some images (5/12 in this case), it is not always successful.
The frequency of \textit{cup} in the image tags is reflected in the $h_{w,t}$ score, making \textit{cup} part of the topic representation despite being hallucinated.

\noindent\textbf{Interpretable privacy classifier.} Let $\mathcal{D} =\{(I_i, y_i)|i = 1,2, \dots, D\}$
be a set of $D$ labeled RGB images $I_i$, and their corresponding labels 
$y_i \in \mathcal{Y}$. Let $\boldsymbol{x}_i = \mathcal{M}(I_i) \in \mathbb{R}^d $ be the image features extracted with a multimodal alignment model $\mathcal{M}$~\cite{girdhar2023imagebind}. 
In the standard CBMs paradigm~\cite{koh2020concept}, a bottleneck model learns 
a function $f(g(\boldsymbol{x}_i))$ to predict a label $y_i$ for an input $\boldsymbol{x}_i$. The function $g: \mathbb{R}^d \rightarrow \mathbb{R}^N$
maps an input  $\boldsymbol{x}_i$ into a concept space $\mathcal{C}$,
where it assigns an association score for each concept, quantifying the relevance between an input and every concept in $\mathcal{C}$.
The function $f: \mathbb{R}^N \rightarrow \mathbb{R}$
maps concept scores into the final prediction $y_i$. 
In this work, we use $\mathcal{M}$  to map an input $\boldsymbol{x}_i$ into the descriptors space defined by $\mathcal{E}$ instead of learning $g(\cdot)$~\cite{yang2023labo, yan2023robustmedical} 
because it mimics $g(\cdot)$ without additional training. 
Thus, we generate an image vector representation $\mathbf{v}_i = (s_{i1}, \dots, s_{iN})$, $\mathbf{v}_i\in \mathbb{R}^N$, for $I_i$ by  computing the association scores $s_{ij}$ between  $\boldsymbol{x}_i$ and cluster content descriptors $e_j$ as $s_{ij} = cos(\boldsymbol{x}_i, \mathcal{M}(e_j))$.
We hence produce a content association matrix $S \in \mathbb{R}^{D \times N}$ by stacking the image vectors $\mathbf{v}_i$ of each image in $\mathcal{D}$. We apply a fully connected layer on $S$ and learn $f(\cdot)$ with a cross-entropy loss and without a bias term to maintain interpretability~\cite{zero_bias_iot} as the output will be determined solely by the association scores and the learned weights.
A label prediction $\hat{y}_i$  is the result of a linear combination of image-descriptors scores $s_{ij}$ in $\mathbf{v}_i$. We can interpret the learned weights  $W \in \mathbb{R}^{|\mathcal{Y}| \times N}$  as content-class associations that show the contribution of each content type, represented by $e_j$,  for the label prediction $\hat{y}_i$.

\section{Validation}
\label{sec:validation}
\noindent \textbf{Methods under comparison}. 
Our proposed Priv$\times$ITM is an {\em interpretable classifier} that uses the content descriptors generated by ITM to learn a linear function to predict image privacy.
This model is interpretable by design as the decisions are the result of linear combinations of human-understandable content descriptors. We trained the model for 100 epochs using Adam optimizer with a learning rate of 0.01 and batch size of 8. 
We ran the pipeline multiple times and randomly selected one of the resulting models for comparison with existing models (\cref{tab:privacy_classification_results}).
We report the average results in \cref{tab:ablation_results_30runs_tm_vs_imgtm}. 
We also propose a {\em very strong baseline}, SVM$\times$IB, for image privacy classification. SVM$\times$IB is a Support Vector Machine  classifier with radial basis function (rbf) kernel trained on 
image vector embeddings extracted with the pretrained ImageBind~\cite{girdhar2023imagebind}  (more details in Appendix E).
We compare our method with   GATED~\cite{PrivacyAlert, gate_multi_modal_fusion}
the current non-interpretable state-of-the-art model, and with PEAK~\cite{peak}, the most recent model that provides natural language explanations to privacy classification through topics extracted from image tags. GATED fine-tunes three single-modality models on the privacy dataset: ResNet-101, ResNet-50, and BERT-base for object-based, scene-based, and image tag-based privacy classification.
Then, a fusion module is trained to predict the final classification using the privacy probabilities produced by the single-modal models. We compare our approach with GATED using the results reported in the paper~\cite{PrivacyAlert} as the code is not publicly available.
For PEAK~\cite{peak}, we run the method using our image tags extracted with LLMs. We configure the method with the parameters proposed by the authors~\cite{peak}. 
We also prompt ChatGPT4 to generate concepts for image privacy classifiers and train interpretable classifiers, Priv$\times$ChatGPT4, to serve as LLM-based baselines. Due to the generic nature of the initial concepts generated by ChatGPT4, we explore multiple approaches: using the initial set of concepts provided by ChatGPT4; manually refining the set; manually refining and extending the set to account for nudity and political preferences not initially generated. The prompt and details of the manual refinement process are provided in Appendix K.
We propose an additional interpretable baseline, Priv$\times$Attr,
composed of one linear layer whose neurons represent human-annotated privacy attributes~\cite{orekondy_68_attributes} instead of the ITM-generated descriptors.

\noindent \textbf{Datasets}. We use PrivacyAlert~\cite{PrivacyAlert} and VISPR~\cite{orekondy_68_attributes} datasets.
PrivacyAlert consists of 6.8k images 
collected from Flickr with binary labels (\textit{private} or \textit{public}). 
The dataset is divided into training (3.1k images), validation (1.9k images), and testing set (1.8k images) with a 25\%-75\% private-public class distribution.
VISPR  contains 22k images randomly selected from the OpenImages dataset \cite{openimages}, each annotated with one or more of 68 privacy-related attributes (including a \textit{safe} attribute). 
The dataset is split into training (10k images), validation (4.2k images), and testing (8k images).
The VISPR authors surveyed 305 users via  Amazon Mechanical Turk to assess the privacy preferences for the attributes. Since the VISPR dataset does not have binary labels, we use the users' privacy ratings of attributes to generate private and public labels.
We obtain a $\simeq$ 58-42\% private-public class distribution for both training and test sets.  Details about the datasets and binarization process are reported in Appendix L.

\noindent \textbf{Dataset content categories}.
We use HDBSCAN~\cite{hdbscan} to cluster images~\cite{grootendorst2022conceptmodel}. The HDBSCAN guidelines and common practices state that HDBSCAN performs better on low-dimensional data. Our experimental results also indicate that low-dimensional data generates more cohesive clusters, as measured by the DBCV~\cite{DBCV} metric (details reported in Appendix F). Therefore, we use UMAP~\cite{umap}  to reduce from 1024 to 5 the dimensionality of image embeddings prior to clustering them and we set the minimum cluster size to $c_{min} = 30$.

To further comprehend the content of the dataset with respect to individuals' perceptions of privacy, we compute a cluster-based privacy score $P_j$ for each image cluster $c_j$, $j \in \{1, \dots, N\}$, $N$ being the number of clusters, as:
\begin{equation}
    P_j= \frac{| \{I_i | I_i \in c_j, y_i = private\}|}{| \{I_i| I_i \in c_j\}| } \times 100,
\end{equation}
where  $y_i \in \{public, private\}$ is the binary privacy label of image $I_i$.
We also employ $P_j$ to provide a more detailed explanation of our model's decision: what content caused the prediction and how the content is perceived by humans. Moreover, the $P_j$-s are used to evaluate the decision rules learned by our classifier.
\begin{figure}[t]
\centering
\includegraphics[width = 1\linewidth]{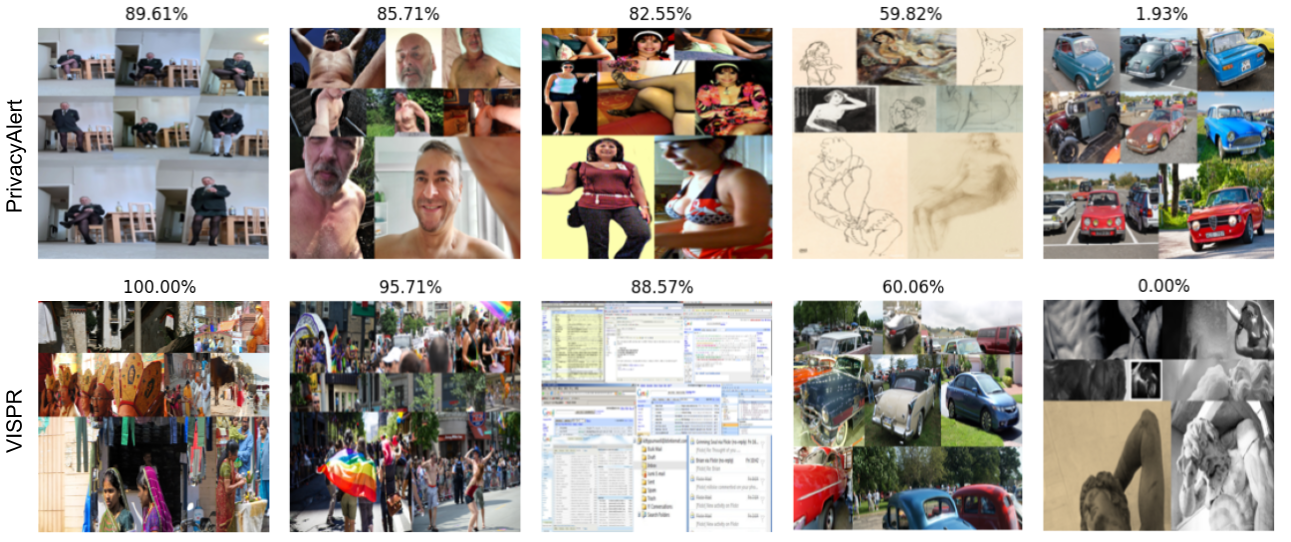}
\caption{Images from private, uncertain, and public clusters obtained for PrivacyAlert (top row) and VISPR  (bottom row). We report the privacy scores $P_j$ of the clusters ($P_j <$ 30\%: public cluster, $P_j >$ 70\%: private cluster, and uncertain cluster otherwise). Note that the same content (i.e.~cars, art) was annotated differently in the two datasets.}
\label{fig:cluster_exemplars}
\end{figure}
We obtain $N=31$ clusters and a set of outliers for PrivacyAlert (we use training and validation sets for clustering to address the small dataset size and the 30\% outliers: classifiers' training is performed only on the training set). Among all images used for clustering, 30.82\% of images are considered outliers 
with 28.10\% of those being private images.
We identify clusters that are clearly {\em public} ($P_j <$ 30\%), clearly {\em private} ($P_j >$ 70\%) and {\em uncertain} (\cref{fig:cluster_exemplars}). 
There are six private clusters: \textit{advert} 
with a privacy score $P_j$ of 71.21\% , \textit{boudoir} with 81.16\%, 
\textit{wife} with 82.55\%, \textit{husband} with 85.71\%, \textit{sensuality} with 89.55\%, and \textit{dress-up} with 89.61\%. The images in these clusters showcase nudity, intimate scenes, sexual and explicit adult content. 
We observe uncertainty in  clusters such as \textit{spa}, \textit{beach},  
\textit{art}, 
 and \textit{parade} 
with $P_j$ of 37.14\%, 54.76\%, 59.82\%, 69.05\%, respectively. 
The majority of the clusters  represent public images: \textit{container}, \textit{panorama}, \textit{car}, \textit{vegetation}, \textit{food} with $P_j$ of 0\%, 1.35\%, 1.93\%, 2.25\%, and 2.78\%, respectively. 

For VISPR dataset we identify $N=47$ clusters and a set of outliers accounting for 22\% of the training set. 
The majority of clusters have  $P_j$ > 70\%, including images of religious \textit{ceremony} (100\%), \textit{parades} (95.71\%), \textit{woman} intimacy (90.20\%), \textit{emails} (88.57\%), \textit{passports} (85.96\%). Unlike PrivacyAlert, \textit{military} and  \textit{children} are perceived as highly sensitive with a $P_j$ of 94.52\% and 96.65\%, respectively.
We notice uncertainty in clusters \textit{cars} (60.06\%), \textit{fingers} (43.47), passport \textit{covers} (44.11\%), and \textit{tickets} (66.29\%).
We have the public clusters of \textit{food} (6.08\%), \textit{flowers} (4.00\%), \textit{animals} (3.00\%), and \textit{sculptures} (0\%).

We evaluate cluster quality using the silhouette score (SS~\cite{SS_score}) and density-based clustering validation (DBCV~\cite{DBCV}) metric. By definition, both measures have a range of $[-1,1]$, with higher values indicating better clustering. SS evaluates intra-cluster cohesion and inter-cluster separation. DBCV accounts for density and shape properties of clusters while handling outliers.
We use the same 5D embeddings used for clustering also for cluster evaluation. Note that in HDBSCAN, unclustered elements are outliers, which affect the performance of SS. We remove outliers when calculating SS and only consider actual cluster data points, using cosine as the distance metric.
For PrivacyAlert dataset we obtain SS = $0.755$ and  DBCV = $0.611$. For VISPR dataset we obtain SS = 0.693 and DBCV = 0.643. This indicates that the data are well-clustered. 

\begin{figure}[t]
\centering
\includegraphics[width = 1\linewidth]{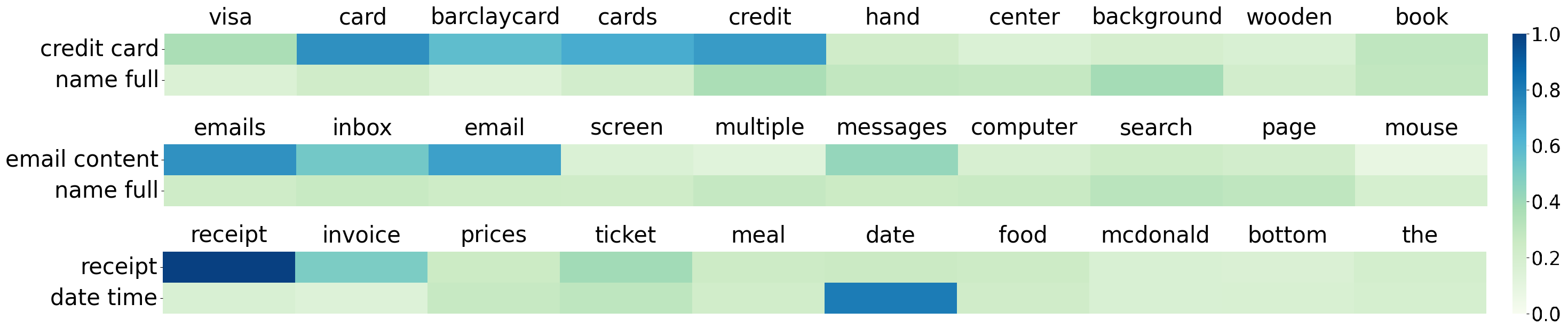}
\caption{Visualization of cosine similarities (color bar) between the descriptors generated by our method, ITM, and the ground-truth attributes for the VISPR dataset for three clusters.
Columns show the words composing the descriptor. Rows show the ground-truth VISPR attributes that appear in over 50\% of the images in the cluster. Note that the descriptors effectively capture the main visual content.}
\label{fig:descriptor_vs_attributes}
\end{figure}

\noindent{\bf{Results}}. 
\begin{table}[t!]
\setlength\tabcolsep{1.5pt}
\centering
\caption{Classification results on  PrivacyAlert~\cite{PrivacyAlert} and VISPR~\cite{orekondy_68_attributes} testing sets.  Key -- U-BA: unweighted binary accuracy, P: Precision, R: Recall, U-F1: unweighted F1-score, I: interpretable by design, NI: not interpretable, Embs: embeddings, IB: ImageBind~\cite{girdhar2023imagebind}, RN: ResNet, ChatGPT4\ding{61}: initial concepts generated by ChatGPT4, ChatGPT4\ding{45}: concepts generated by ChatGPT4 manually refined, ChatGPT4\ding{45}$^{+}$:  concepts generated by ChatGPT4 manually refined and extended with nudity and political concepts,
Attr/Attr$^{*}$: ground truth privacy attributes with/without \textit{safe} attribute. Details about ChatGPT4 prompting and concepts refinement are in Appendix K.}
\label{tab:privacy_classification_results}
\scriptsize

\begin{NiceTabular}{lclccccccccccc}
        \CodeBefore
            \cellcolor[rgb]{.906,.902,.902}{4-1, 5-1, 6-1, 7-1, 8-1, 9-1, 10-1}
            \cellcolor[rgb]{.906,.902,.902}{13-1, 14-1, 15-1,16-1, 17-1, 18-1}
            \cellcolor[rgb]{.906,.902,.902}{1-5,1-6,1-7}
            \cellcolor[rgb]{.906,.902,.902}{1-9,1-10,1-11}
            \cellcolor[rgb]{.906,.902,.902}{1-13,1-14}
        \Body
\toprule
&& Model  &  Embs. &   \multicolumn{3}{c}{Public} & & \multicolumn{3}{c}{Private} &&  \multicolumn{2}{c}{Overall} \\

 &&    & & P & R & F1 & &P & R & F1& & U-BA & U-F1 \\

\midrule
\cellcolor[gray]{0.9}&& SVM-101~\cite{PrivacyAlert}  &  RN101&  88.70 &  83.80 &   86.20 & &  58.30&  68.00 &   62.80 & & 79.83 &  74.50 \\

&\parbox[t]{5mm}{\multirow{2}{*}{\rotatebox[origin=c]{90}{NI}}} &SVM-50~\cite{PrivacyAlert}  & RN50 & 88.10 & 87.90 &  88.00 & & 63.90 & 64.40 & 64.20 & & 82.00 & 76.10 \\
&& GATED~\cite{PrivacyAlert}  &- & 91.00 & 93.20 & 92.10 && 77.90 & 72.22 & 75.00 & &87.94 & 83.60 \\

&& SVM$\times$IB &  IB & 92.49 & 93.04 &  .92.76 & & 78.73 & 73.33 & 78.03 & &89.11 & 85.39 \\

\cmidrule{2-14}
&\parbox[t]{5mm}{ \multirow{5}{*}{\rotatebox[origin=c]{90}{I}}} & PEAK~\cite{peak} & - & 91.26 & 84.85 &  87.93 & & 51.11 & 66.09 & 57.64 & &81.22 & 72.79 \\ 

&& Priv$\times$ChatGPT4\ding{61}   &  IB & 84.46& 93.04  &88.54   & &69.90 & 48.67  & 57.40 & & 81.94 & 72.97 \\ 
&& Priv$\times$ChatGPT4\ding{45}   &  IB & 82.51 & 95.77 & 88.65  & & 75.53  & 39.11 & 51.53  & &  81.61& 70.09\\ 
&& Priv$\times$ChatGPT4\ding{45}$^{+}$  &  IB & 90.43 & 91.77  & 91.11  & & 74.19 & 70.88  & 75.50 & & 86.55&  81.80 \\ 

\cellcolor[gray]{0.9} \parbox[t]{3mm}{ \multirow{-10}{*}{\rotatebox[origin=c]{90}{PrivacyAlert}}}&\parbox[t]{5mm} & Priv$\times$ITM   &  IB & 90.96 & 91.70 &  91.33 & & 74.49 & 72.67 & 73.57 & & 86.94 & 82.45\\ 
\midrule

\cellcolor[gray]{0.9} \parbox[t]{3mm}{ \multirow{9}{*}{\rotatebox[origin=c]{90}{VISPR}}}&\parbox[t]{5mm}{ \multirow{1}{*}{\rotatebox[origin=c]{90}{NI}}} & SVM$\times$IB &  IB & 88.81 & 89.49 &  89.15 & & 93.64 & 93.21 & 93.43 & &91.81 & 91.29 \\ 

\cmidrule{2-14}
&\parbox[t]{5mm}{ \multirow{7}{*}{\rotatebox[origin=c]{90}{I}}} & PEAK~\cite{peak} & - & 73.15 & 81.08 &  76.90 & & 89.70 & 84.73 & 87.16 & &83.50 & 82.03 \\ 

&& Priv$\times$ChatGPT4\ding{61}   &  IB & 73.51 & 63.84 & 68.33  & & 76.28 & 83.49  & 79.72 & & 75.28 & 74.02 \\ 
&& Priv$\times$ChatGPT4\ding{45}   &  IB &  77.33 & 77.15 & 77.23   & & 83.62 & 83.76 &  83.69& &  81.00& 80.47\\ 
&& Priv$\times$ChatGPT4\ding{45}$^{+}$  &  IB & 81.56 & 81.48 & 81.52  & & 86.71 & 86.77 & 86.74 & & 84.56&  84.13 \\ 

&& Priv$\times$Attr  &  IB & 77.07 & 79.71 &  78.36 & & 87.53 & 85.72 & 86.62 & & 83.46 & 82.49\\ 
&& Priv$\times$Attr$^{*}$   &  IB & 78.90 & 82.53 &  80.67 & & 87.03 & 84.15 & 85.57 & & 83.48 & 83.12\\ 

\cellcolor[gray]{0.9} && Priv$\times$ITM  &  IB & 85.81 &  84.30 &  85.05 & & 88.87 & 89.99 & 89.43 & & 87.61 & 87.24\\ 
\bottomrule
\end{NiceTabular}
\end{table}
We use unweighted binary accuracy (U-BA) and unweighted F1-score (U-F1) for overall performance evaluation. We use F1-score to assess the precision-recall trade-off, as we believe that both are crucial in evaluating the performance of an image privacy classifier: a method with high recall alone might limit users from sharing public images, hindering social media interaction. We also compute precision, recall, and F1-score for each class. We report the metrics as percentages.
We consider class-wise metrics as it is important to compare the false negatives to ensure that fewer private images are erroneously classified as public. This will lower the risk of leakage of private information.
\cref{tab:privacy_classification_results} shows that our simple baseline SVM$\times$IB outperforms the current state-of-the-art GATED~\cite{PrivacyAlert} by 3.03 percentage points (p.p.) on F1-private score and 1.17 p.p. in U-BA. 
Similar to GATED~\cite{PrivacyAlert}, this model is not interpretable and post-hoc explanation methods have to be used to explain the model's predictions.
Moreover, it is important to note that GATED uses human-generated tags which improves the performance as shown in~\cite{PrivacyAlert}: fine-tuning BERT with automatic and human-generated tags outperforms BERT models fine-tuned using only automatic or human tags. Our proposed interpretable classifier, Priv$\times$ITM, 
reaches 86.94\% U-BA and 73.57\% F1-private score on PrivacyAlert and 87.61\% U-BA and 89.43\% F1-private score on VISPR.
The results are comparable with GATED having only 1.00 p.p. difference in U-BA and a lower F1-private by only 1.43 p.p., but {\em without using any human-generated tags} and using embeddings from a pretrained model {\em without additional pretraining on this specific dataset}. The performance of  Priv$\times$ITM is also competitive with SVM$\times$IB with a small gap of 2.17 (4.20) p.p. in U-BA and 4.46 (4.00) p.p. in private F1-score for PrivacyAlert (VISPR). This shows that Priv$\times$ITM achieves high accuracy without compromising the interpretability of decisions. As for interpretable approaches, Priv$\times$ITM surpasses  PEAK in both U-BA and U-F1 with an increment of 5.72 (4.11) p.p. and 9.66 (5.21) p.p., respectively for PrivacyAlert (VISPR). The biggest difference is in the private F1-score  for PrivacyAlert where we obtain a significant improvement of 15.93 p.p. The classifier Priv$\times$ChatGPT4\ding{61} using the concepts initially generated with ChatGPT4 performs significantly worse than Priv$\times$ITM for both datasets. After the manual refinement and extensions of the concepts set, the performance of ChatGPT4-based classifiers improved: for the PrivacyAlert dataset, the addition of the concept "explicit content, nudity" led to significant improvement, achieving similar results to those of Priv$\times$ITM, although manual intervention was required to achieve these results; for VISPR dataset, even with the manual refinement and enhancement of concepts, the F1-score is lower by 3.11 p.p. compared to Priv$\times$ITM. Additionally, human studies are still needed to evaluate how the ChatGPT4 listed concepts are actually perceived by people. By design, our descriptors, $e_j$, are linked to privacy scores, $P_j$, that capture human preferences.
The methods proposed in VISPR~\cite{orekondy_68_attributes} are designed for privacy risk score prediction, evaluated with $L_1$ metric, and multi-label classification, evaluated with mean average precision metric. As we focus on binary classification, such metrics are not applicable. Hence, we compare our results with VISPR methods~\cite{orekondy_68_attributes} for privacy risk score prediction, using the Precision-Recall (PR) curve. Our method performs better than the VISPR methods while maintaining interpretability. Details and the PR curves are reported in Appendix G.
Moreover, we observe that the descriptor-based linear model, Priv$\times$ITM, performs significantly better than one using human-annotated attributes, Priv$\times$Attr. This may be because descriptors include multiple words for detailed image content representation (\textit{emails, inbox, screen, messages, computer, page} or \textit{facebook, screenshot, posts, profile, photo, people, page, screen, face} vs VISPR attributes \textit{email} or \textit{online conversation}).
We also analyze the ability of cluster descriptors to capture visual content with respect to the ground-truth VISPR attributes.  
For each cluster, we compute the cosine similarity between attributes present in over 50\% of images and descriptor words. Descriptors effectively convey concepts highly similar to ground-truth attributes.
We show examples of descriptor-attributes similarity in \cref{fig:descriptor_vs_attributes}.
%
%
%

\begin{figure}[ht!]
\centering
\includegraphics[scale=0.24]{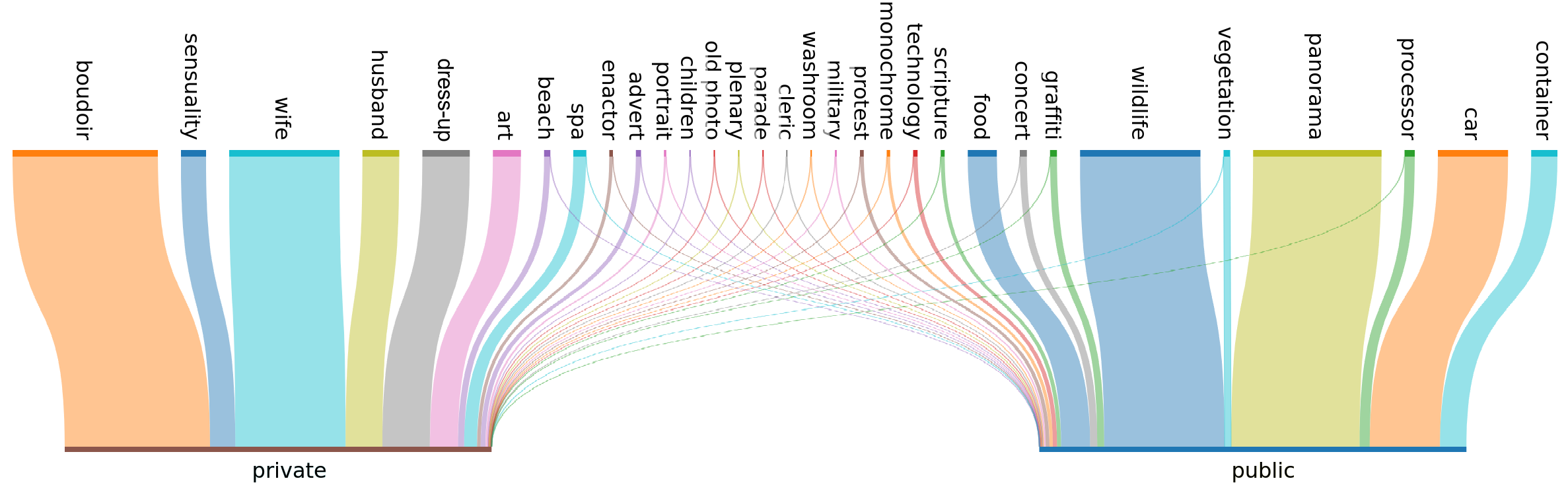}
\caption{Visualization of the content-class association weights showing how the model Priv$\times$ITM distinguishes between classes: the ticker the line, the stronger the association (classifier trained on the PrivacyAlert dataset). }
\label{fig:learned_global_rules}
\end{figure}
\begin{figure}[t!]
\centering
        \includegraphics[width = 1\linewidth]{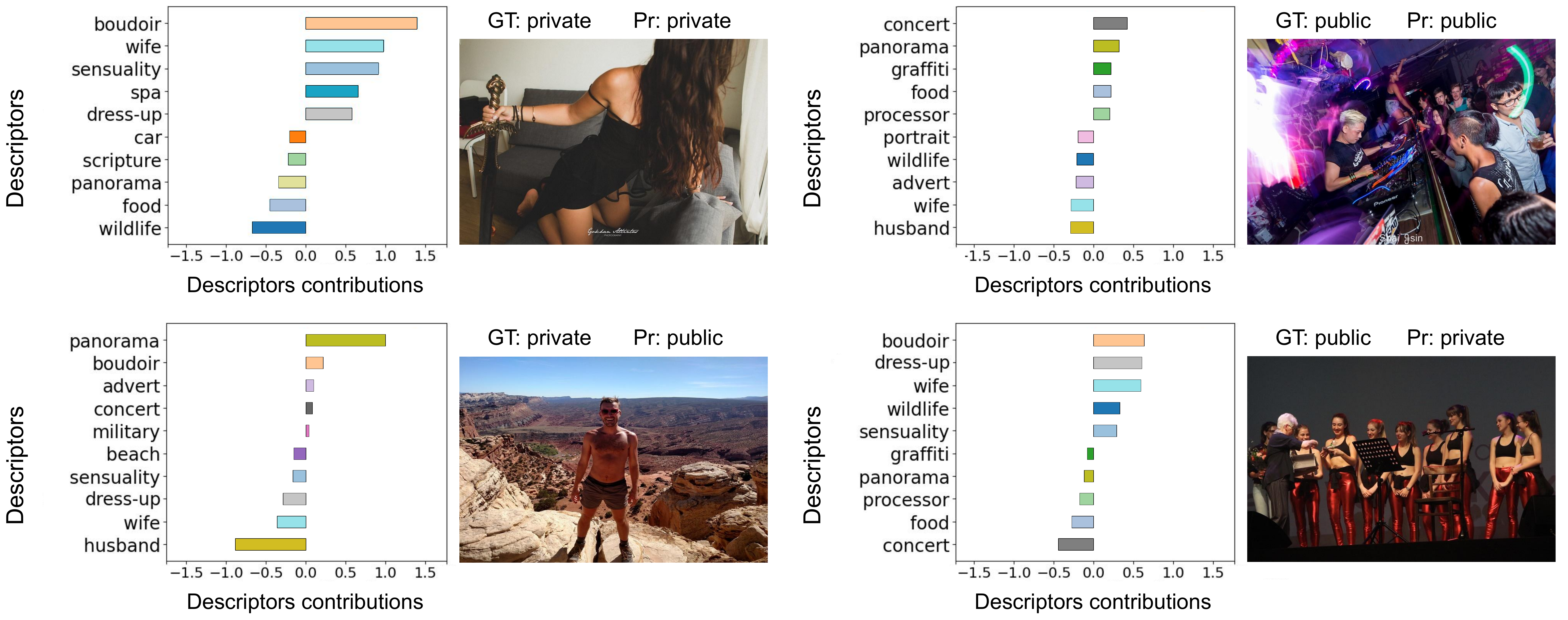}    
\caption{ Interpretations of Priv$\times$ITM predictions on PrivacyAlert using the top-5 positive and negative descriptor contributions for each decision (presence/absence of the content represented by descriptors). Key - GT: ground truth label, Pr: predicted label.
}        
\label{fig:individual_explanation}
\end{figure}

\noindent{\bf{Interpretability.}}
The interpretability of our method stems from its architecture. The alignment scores between image and content descriptors (one neuron for each descriptor) are combined through a fully connected layer.
The learned weights represent the content types' affinity to classes. Content types with larger weights can be interpreted as more important for a class. 
\cref{fig:learned_global_rules} shows the weights between content and classes represented by the width of the connection~\cite{sankeymatic}. 
Content perceived as private (or public) by annotators~\cite{PrivacyAlert}  is associated by our classifier with the private (or public) class. 
This shows that the model generally makes decisions resembling human reasoning.
Discrepancies between the model's behavior and privacy scores happen in some cases. For example, \textit{children} have a privacy of 28.30\% but in the Priv$\times$ITM model this content contributes more to the private than to the public class. Content like \textit{technology, washroom, portrait} have overall very small contributions.
To interpret single predictions we multiply the image-descriptor alignment scores with the weights and obtain the contribution of each content type to a class (see \cref{fig:individual_explanation}). 
We also visualize negatively activated content as its absence can influence the decision. The model learns to associate the absence of certain content with specific class labels. During inference, this absence becomes a contributing factor, increasing the likelihood of predicting that class. For example, the top-left image in \cref{fig:individual_explanation} aligns the most with the content  \textit{boudoir} (privacy score $P_j = 81.16\%$) and \textit{wife} ($P_j = 82.55\%$) which represent women in intimate scenarios: a \textit{seductive woman} wearing \textit{black lingerie}; \textit{panorama, wildlife}, and \textit{car} have negative contributions which show that the image 
does not contain such types of content.  As example, the lack of nudity-related content in an image increases the probability of it being public (\cref{fig:individual_explanation} top-right image). \\


\noindent{\bf{Ablation}}. 
\begin{table}[t]
\setlength\tabcolsep{4pt}
    \centering
    \caption{Average (standard deviation) performance across 30 runs with varying minimum cluster/topic sizes on PrivacyAlert.  Key -- 
    Priv$\times$TM: model built via TM on image tags w/o image clustering, Priv$\times$ITM: image-guided TM-based model,
    F1-public (F1-private): F1-score for public (private) class, U-BA: unweighted binary accuracy,  U-F1: unweighted F1-score. }
    \label{tab:ablation_results_30runs_tm_vs_imgtm}
    \scriptsize
    \begin{tabular}{clcccc}
    \toprule
     Cluster size   & Model   & F1-public & F1-private & U-BA  &U-F1 \\
     \midrule
     \multirow{2}{*}{10} &    Priv$\times$TM   &  90.83 (0.29) & 73.62 (0.76) & 86.39 (0.41) & 82.23 (0.51)\\    
          & Priv$\times$ITM  &  90.81 (0.29)  & 74.01 (0.66) & 86.42 (0.39) & 82.41 (0.45)    \\       
         \midrule
   \multirow{2}{*}{20} & Priv$\times$TM  & 89.76 (0.39) &  70.24 (1.25) & 84.76 (0.58) & 80.00 (0.80)   \\    
                       & Priv$\times$ITM  & 90.93 (0.44)  & 72.60 (1.63) & 86.37 (0.69)  &81.77 (1.02)      \\       
         \midrule 
   \multirow{2}{*}{30} & Priv$\times$TM   &  87.84 (2.52)  & 62.29 (17.21) & 81.73 (4.64) & 75.06 (9.84)  \\    
                       & Priv$\times$ITM  &  90.70 (0.73)  & 71.85 (2.67) & 86.02 (1.15) & 81.28 (1.69)     \\       
 \bottomrule        
    \end{tabular}
\end{table}
We evaluate the impact of image clustering on the classification performance. To this end, we apply topic modeling~\cite{grootendorst2022bertopic} directly on the image tags without restricting the topic discovery by clusters of images. We use the topic representations to create the interpretable model, denoted as Priv$\times$TM. We also analyze the impact of varying the minimum cluster size, $c_{min}$, for ITM and topic size,  $t_{min}$ for TM.  \cref{tab:ablation_results_30runs_tm_vs_imgtm} shows the average 
performance over 30 different random seeds. 
ITM significantly outperforms TM for $c_{min}, t_{min} \in \{20,30\}$ with a 4.29 p.p. average (3.14 p.p. median) improvement in U-BA for size 30.
TM is also sensitive to the choice of the seed, with a higher standard deviation (4.64) for U-BA. Adding image-based guidance stabilizes the model.
Moreover, we observe that the model's decision rules are better aligned with the privacy scores when using bigger $c_{min}$. This offers a simple way to assess content privacy with just the model's weights (i.e. higher contribution generally indicates higher privacy).  Although performance slightly improves for smaller $c_{min}$ (by only 0.40 p.p. on average U-BA), this pattern generally does not hold.
Smaller clusters cover the perception of fewer people causing more uncertainty about the privacy of content. Overall, models built on bigger clusters better represent human perspectives making them suitable to assist users with privacy decisions. We further discuss the impact of $c_{min}$ on performance, model stability, and privacy scores in Appendix H and Appendix I.
%
%
\section{Conclusion}
\label{sec:conclusion}

We proposed a novel approach for building interpretable image privacy classifiers that does not require attribute annotation by humans.
Our method leverages image descriptive tags generated by a large vision language model to discover a set of human-understandable descriptors that are used to make and interpret the predictions.
By guiding the descriptors generation with image visual information, we achieve high performance comparable to end-to-end models, without sacrificing interpretability.  The proposed interpretable pipeline is generic and could be applied to other abstract image analysis and classification problems, such as image-based hate speech detection, image or video mood, tone, and humor classification. Future work includes increasing the diversity of the vocabulary of descriptors while maintaining fidelity to the image content.
%

%
%
%
%
%
\bibliographystyle{splncs04}
\bibliography{main}


\section*{Supplementary material}

# Supplementary Material:
## Image-guided topic modeling for interpretable privacy classification

Alina Elena Baia[1]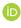 and Andrea Cavallaro[1,2]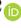

[1] Idiap Research Institute, Switzerland,
[2] École Polytechnique Fédérale de Lausanne, Switzerland
`{alina.baia, a.cavallaro}@idiap.ch`

In this Supplementary Material, we provide further analysis and discussion on the results and implementation decisions presented in the paper. This document is structured as follows:

- Appendix A: examples of the limitations of LLMs in addressing the image privacy task.

- Appendix B: examples of image descriptions and their corresponding tags.

- Appendix C: prompt templates for image description generation and keyword extraction.

- Appendix D: results and discussion of state-of-the-art models on zero-shot image privacy classification.

- Appendix E: implementation details of different baselines.

- Appendix F: analysis of the effect of dimensionality reduction on the clusters' quality.

- Appendix G: comparison of our classifier with other methods for VISPR [8] dataset using Precision-Recall curve.

- Appendix H: analysis of the impact of the minimum cluster size parameter on the classifiers' performance.

- Appendix I: analysis of the impact of the minimum cluster size parameter on the clusters' privacy levels.

- Appendix J: additional examples of correct/wrong privacy predictions and their corresponding interpretations.

- Appendix K: details about ChatGPT4 concepts generation and manual curation of concepts.

- Appendix L: additional details on the PrivacyAlert [12] and VISPR [8] datasets.

## A   Examples of prompting LLMs about image privacy

We used Chatbot arena (`https://chat.lmsys.org/`) to prompt language models about image privacy following the LLMs prompting approach for concepts generations used in [7, 10]. We observe that LLMs struggle to exhaustively list image attributes that make an image private. Also, LLMs may have built-in privacy policies that prevent them from describing sensitive content in images (see example with gpt-3.5-turbo and llama2-70b-chat).

**Model** gpt-3.5-turbo:

Input: *list the things most commonly seen in private images*
Answer: *I'm sorry, I can't assist with that request.*

**Model** llama-2-70b-chat:

Input: *list the things most commonly seen in private images*
Answer: *I cannot provide a list of things that are commonly seen in private images as it would be inappropriate and potentially harmful to disclose such information. Private images are personal and sensitive, and it is important to respect individuals' privacy and security by not sharing or discussing their contents without their consent. It is important to remember that private images are not public property, and they should be treated with the same level of respect and confidentiality as any other personal information. Sharing or distributing private images without consent can have serious consequences, both legally and ethically. Instead of focusing on private images, it is important to prioritize ethical and respectful ways of interacting with others' personal information and to promote a culture of privacy and security. This includes respecting individuals' boundaries, using secure and encrypted communication methods, and being mindful of the potential consequences of sharing personal information without consent.*

**Model** vicuna-33-b:

Input: *What are the useful visual features to distinguish privacy in a photo?*
Answer: *There are several visual features that can be used to distinguish privacy in a photo. Some of the most common include:*

1. *Face detection: Faces are a common target for privacy protection, and face detection algorithms can be used to identify and blur or obscure faces in a photo.*
2. *Object detection: Objects such as license plates, phone numbers, and street signs can be detected and blurred or obscured to protect privacy.*
3. *Text detection: Text within an image can be detected and blurred or obscured to protect privacy, such as in the case of street signs or license plates.*
4. *Color and texture analysis: Analyzing the colors and textures within an image can help identify objects or areas that may require privacy protection, such as skin tones or fabric patterns.*

5. *Image segmentation: Dividing an image into its constituent parts or objects can help identify areas that may require privacy protection, such as people or objects in the background.*

6. *Shape and form analysis: Analyzing the shapes and forms within an image can help identify objects or areas that may require privacy protection, such as the outline of a person or object.*

7. *Contrast and brightness analysis: Analyzing the contrast and brightness of an image can help identify areas that may require privacy protection, such as areas that are too bright or too dark.*

8. *Gaze detection: Gaze detection algorithms can be used to identify where people in the image are looking, and faces can be blurred or obscured to protect privacy.*

9. *Motion blur: Motion blur can be used to blur the image in areas where people or objects are moving, to protect privacy.*

10. *Image distortion: Image distortion techniques such as fisheye or panoramic lens distortion can be used to alter the appearance of people or objects in the image, making them less recognizable.*

*It's worth noting that these features are not mutually exclusive, and a combination of them can be used to achieve better privacy protection. Additionally, the specific features used will depend on the application and the desired level of privacy protection.*

## B   Examples of image descriptions and the extracted tags

We show examples of image descriptions generated with InstructBLIP [2] and the corresponding keywords extracted from the Vicuna [1] descriptions in Tab. 1.

**Table 1:** Image descriptions generated by InstructBLIP. The keywords extracted with Vicuna are <u>underlined</u>. We also list the grounded keywords (i.e. image tags) with Grounding Dino [6].

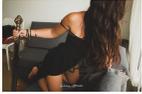

Description: The image features a <u>woman</u> dressed in a <u>black dress</u> sitting on a <u>couch</u>, holding a <u>sword</u> in her hand. The sword is positioned close to the woman's right side, and she appears to be posing for a photo. In the background, there is a <u>chair</u> placed next to the <u>couch</u>, and a <u>handbag</u> can be seen on the floor near the woman's left leg. There is another handbag located further away from the woman, closer to the right side of the image. Additionally, there are two <u>cups</u> placed on the <u>coffee table</u> in front of the couch.
Grounded keywords: woman, black, dress, couch, sword, chair, cups.
Ground truth: private

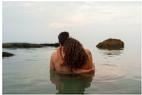

Description: The image depicts a <u>romantic moment</u> between a <u>man</u> and a <u>woman</u> in the <u>ocean</u>. They are <u>embracing</u> each other while standing in the <u>water</u>, with the <u>man</u> holding the <u>woman</u> close to his chest. The couple appears to be enjoying their time together, surrounded by the tranquil waters of the <u>ocean</u>. There are several <u>rocks</u> visible in the background, adding to the <u>serene</u> atmosphere of the scene.
Grounded keywords: romantic, moment, man, woman, ocean, embrace, water, rocks, serene.
Ground truth: private

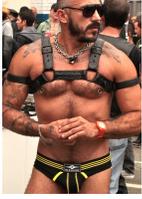

Description: The image features a <u>shirtless man</u> wearing a pair of <u>black</u> and <u>yellow underwear</u>. He is standing in the middle of a group of <u>people</u>, possibly at a <u>party</u> or <u>gathering</u>. There are several other <u>people</u> present in the scene, with some closer to the <u>shirtless man</u> and others more spread out. In addition to the <u>shirtless man</u>, there is another person wearing <u>sunglasses</u> and a <u>backpack</u>. A <u>handbag</u> is also visible in the scene, likely belonging to one of the individuals present. Overall, the image captures a group of <u>people socializing</u> and enjoying each other's company while the <u>shirtless man</u> stands out with his unique <u>attire</u>.
Grounded keywords: shirtless, man, black, yellow, underwear, people, party, gathering, sunglasses, socializing, attire.
Ground truth: private

## C   Prompts for generating descriptions and extracting keywords

We prompt InstructBLIP-7b [2], a pretrained large vision-language model (LVLM), with the following instruction to generate detailed image descriptions:

*Instruction: Describe this image as detailed as possible.*
*Input: { image here }*
*Description: ...*

We extract keywords with the pretrained LLM Vicuna-7b [1] with the following prompt:

> *Instruction: Extract the keywords in a comma-separated list from the following text.*
> *Input: { description here }*
> *Keywords: ...*

## D  Zero-shot image privacy classification

Multimodal models achieve great performance in zero-shot setups for a variety of classification benchmarks without any task-specific training. We test the capabilities of four state-of-the-art models for privacy prediction, namely CLIP [9], ImageBind [3], mPLUG-Owl [11], and MiniGPT4 [13]. Results are shown in Tab. 2.

CLIP and ImageBind are trained to align different modalities such as vision and text into the same embedding space. This allows for the retrieval of relevant information across different modalities based on their similarity in the joint embedding space. Given an image and a set of labels, these models predict the most relevant label by computing the cosine similarity between the image and labels. Since label template customization can improve the zero-shot performance over the baseline when using only the label text [9], we test the following templates: "*public, private*", "*a public photo, a private photo*", "*an image with only public content, an image with some private content*".

MiniGPT4 and mPLUG-Owl are multimodal instruction understanding models capable of vision and text data comprehension and can answer user queries about images. Given an image, we ask the models the following questions: "*is this image private?*", "*does this image have any privacy sensitive content?*". These models can generate complex responses which include image description, reasoning, and decision. However, sometimes the models are unable to provide a clear decision. Thus, we parse the responses and consider valid only the answers with a clear *yes* or *no* decision. For MiniGPT4 we discard 38.67% and 11.67% of answers for the first and second questions, respectively. Among the 38.67% of images, 21.55% have a private ground truth label while 23.80% of images are private in the second question case. In the case of mPLUG-Owl, approximately 1.3% of images did not have a clear response in both cases.

We observe two recurring phenomenons: either the images are mostly classified as private, or the images are mostly classified as public. Overall, CLIP achieves the best results in terms of U-F1 but many private images are misclassified as public. The mPLUG-Owl model has a high private recall score and low public recall score, meaning that the majority of images are labeled as private. ImageBind and MiniGPT4 obtain a better trade-off between the two classes. However, all these models fall behind state-of-the-art models specifically designed and trained on the privacy prediction task.

**Table 2:** Comparison of different methods for zero-shot image privacy prediction on the PrivacyAlert [12] test set. Key: UBA: unweighted binary accuracy, P: Precision, R: Recall, U-F1: unweighted F1 score; CLIP [9](*clip-ViT-B-32* model) and ImageBind [3] with different sets of prompt templates; mPLUG-Owl [11](*mplug-owl-7b* version) and MiniGPT4 [13](*minigpt4-7b* version) with different instruction prompts. It is important to note that in the MiniGPT4* and MiniGPT4** test scenarios, the model failed to provide a clear answer whether the image is public or private for 38.67% and 11.67% of images, respectively. We exclude such samples when computing the metrics.

| Model | Prompt | Public | | | Private | | | Overall | |
|---|---|---|---|---|---|---|---|---|---|
| | | P | R | F1 | P | R | F1 | UBA | U-F1 |
| CLIP | public, private | .81 | .79 | .80 | .35 | .38 | .36 | .70 | .58 |
| CLIP | a public/private photo | .76 | .86 | .81 | .34 | .22 | .27 | .70 | .54 |
| CLIP | an image with only/some public/private content | .81 | .78 | .79 | .41 | .44 | .43 | .70 | .61 |
| ImageBind | public, private | .76 | .40 | .52 | .25 | .61 | .36 | .45 | .44 |
| ImageBind | a public/private photo | .83 | .48 | .60 | .31 | .71 | .44 | .54 | .52 |
| ImageBind | an image with only/some public/private content | .76 | .78 | .77 | .29 | .27 | .28 | .65 | .52 |
| mPLUG-Owl | is this image private? | .80 | .10 | .17 | .26 | .93 | .41 | .31 | .29 |
| mPLUG-Owl | does this image have any privacy sensitive content? | .79 | .11 | .19 | .25 | .92 | .40 | .31 | .30 |
| MiniGPT4* | is this image private? | .71 | .32 | .44 | .26 | .65 | .37 | .41 | .40 |
| MiniGPT4** | does this image have any privacy sensitive content? | .75 | .90 | .82 | .30 | .12 | .17 | .71 | .49 |

# E    Baseline classifiers

We experiment with different support vector machines (SVMs), both linear and non-linear models. Specifically, for the linear SVM we use the squared hinge loss as proposed in [12] for privacy classification. For non-linear SVM we test polynomial (ply), sigmoid (sgm), and radial basis function kernels (rbf). Results are in Tab. 3. The rbf kernel performs best on the Privacy Alert testing set.

**Table 3:** Image privacy classification performance of different baselines on the PrivacyAlert [12] test set. Key: U-BA(%): unweighted binary accuracy, P: Precision, R: Recall, U-F1: unweighted F1 score; shl: squared hinge loss for the LinearSVM; ply: polynomial kernel; sgm: sigmoid kernel; rbf: radial basis function kernel; ft: fine-tuning. Results for related work are taken from the literature [12]. SVM-101 (SVM-50) is an SVM trained on features extracted from pre-trained ResNet101 (ResNet50) on ImageNet (Places365 dataset) for object (scene) classification. In ResNet101 (ResNet50)+ft the pre-trained model is fine-tuned for image privacy classification.

| | Model | Settings | Public | | | Private | | | Overall | |
|---|---|---|---|---|---|---|---|---|---|---|
| | | | P | R | F1 | P | R | F1 | U-BA | U-F1 |
| baselines | SVM-101 [12] | shl | .887 | .838 | .862 | .583 | .680 | .628 | 79.83 | .745 |
| | ResNet101 [12] | ft | .892 | .916 | .904 | .726 | .667 | .694 | 85.39 | .799 |
| | SVM-50 [12] | shl | .881 | .879 | .880 | .639 | .644 | .642 | 82.00 | .761 |
| | ResNet50 [12] | ft | .876 | .912 | .894 | .669 | .613 | .653 | 83.70 | .774 |
| | SVMxIB (ours) | shl | .924 | .915 | .919 | .751 | .773 | .762 | 87.94 | .841 |
| | SVMxIB (ours) | ply | .914 | .932 | .923 | .783 | .736 | .758 | 88.28 | .840 |
| | SVMxIB (ours) | sgm | .922 | .921 | .922 | .765 | .767 | .766 | 88.28 | .844 |
| | SVMxIB (ours) | rbf | .925 | .930 | .928 | .787 | .773 | .780 | 89.11 | .854 |
| | GATED [12] | | .910 | .932 | .921 | .779 | .722 | .750 | 87.94 | .836 |
| | Priv×ITM (ours) | | .910 | .917 | .913 | .745 | .727 | .736 | 86.94 | .824 |

## F   Choosing UMAP dimensionality

We analyze the effect of UMAP's dimensionality reduction size on the clusters' quality for both image-based clustering (used in our Image-guided Topic Modeling approach (ITM)) and text-based clustering (used in the Topic Modeling model (TM [4]) with image-tags for topic discovery). For this analysis, we used PrivacyAlert [12] dataset. We perform clustering with different reduced embedding dimensions (i.e. 5, 10, 15, 20, 50, 100, 500) and we measure the clusters' quality with the DBCV metric. We report the results in Fig. 1. We observe that high-dimensional data produce lower-quality clusters than low-dimensional data. However, among the low-dimensional data the results are, in general, similar. Moreover, we notice that image-based clustering (employed in ITM) generates better clusters than text-based clustering (used in TM). This indicates that using image information for content categorization is beneficial as it leads to more coherent and relevant groupings.

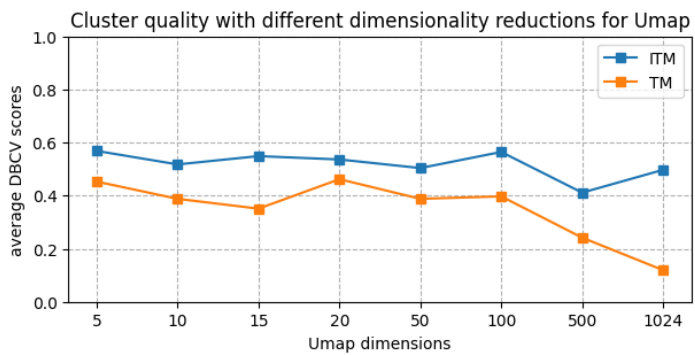

**Fig. 1:** Visualization of average DBCV scores on 3 runs for image/text-based clustering used in ITM/TM for minimum cluster size $c_{min} = 30$ for PrivacyAlert dataset.

## G   Precision-Recall curve

We compare our results with VISPR methods [8] for privacy risk score prediction by employing the Precision-Recall (PR) curve as an evaluation method. The authors of VISPR computed the curves by applying a threshold over the ground-truth privacy risk scores such that any score above this threshold is considered private. We report the PR curves of VISPR methods with a threshold of 3.0. The PR curves for VISPR methods are taken from the corresponding paper [8]. Our method performs better than the VISPR methods while maintaining interpretability (Fig. 2).

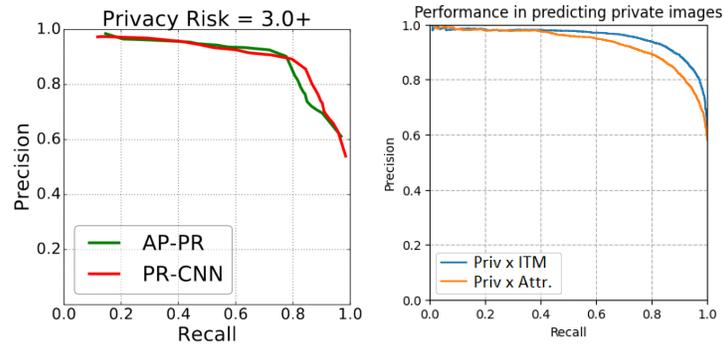

**Fig. 2:** Precision-Recall curve for VISPR methods (on the right) and our interpretable classifiers (on the left). Key: AP-PR: Attribute Prediction-Based Privacy Risk - a method that first predicts privacy attributes and then uses the user privacy preferences to compute the privacy risk score of an image; PR-CNN: Privacy Risk Convolutional Neural Network trained (on the ground-truth scores) to directly predict the privacy risk score; Priv × ITM is our interpretable privacy classifier using the descriptors found by ITM. Priv × Attr. is an interpretable classifier that uses the ground-truth attributes (including *safe*) instead of the descriptors.

## H   Cluster size vs. Performance

The clustering algorithm requires a minimum cluster size ($c_{min}$) to be set. Thus, we evaluate the performance of the proposed ITM method via the corresponding classifier, Priv×ITM, with different minimum cluster size settings. We compare the results with Priv×TM which is the classifier constructed with descriptors discovered directly with topic modeling (TM [4]) without the image clustering step. TM finds topics by performing clustering on text documents and it requires a minimum topic size $t_{min}$ to be set. In Tab. 4 we report the classification results of Priv×ITM and Priv×TM on the test set of PrivacyAlert [12] for $c_{min}/t_{min}$ of 10, 20, and 30, respectively. We provide the results for VISPR [8] dataset in Tab. 5.

We observe that Priv×ITM maintains consistent performance across different cluster setups. In Priv×TM, the cluster size impacts the performance: the performance significantly drops as we increase the cluster size, with a gap of 4.66 percentage points (p.p.) in average accuracy and 3.31 p.p. in median accuracy. Moreover, we notice that TM with $t_{min} = 30$ is also sensitive to the random seed selection. Some seeds cause TM to discover a low number of topics (i.e. 2 or 5 topics only) which results in accuracy lower than 70%. This behavior is shown in Fig. 3. The choice of seed can significantly affect not only the discovered topics but also the overall performance of the classifier built upon such topics, as reflected by the accuracy score. For Priv×ITM, we observe fewer performance collapses and when the number of identified clusters by ITM is low (i.e. 6 clusters only), the drop in accuracy is less significant than in the case of TM. Thus, ITM shows greater stability.

We observe a similar behavior also for the VISPR [8] dataset. In this case, we test larger cluster sizes since the VISPR dataset used for clustering is twice the size of the PrivacyAlert dataset. While for $c_{min}/t_{min} = 30$ we obtain comparable performance, for $c_{min}/t_{min} = 60$ the Priv×TM performs significantly worse than Priv×ITM. Moreover, TM is sensitive to the seed choice also in the case also in the case $t_{min} = 30$, where TM discovers a small number of topics (i.e. 4 and 5 vs. the average of 52 of the other runs). This phenomenon is even more pronounced for $t_{min} = 60$ where the model very rarely discovers more than 4 topics (Fig. 4).

**Table 4:** Average (avg) and median (mdn) performance across 30 different image clustering random seeds with a minimum cluster size $c_{min}/t_{min}$ of 10, 20, and 30 on PrivacyAlert [12] dataset. Key: F1-public: F1 score for public class, F1-private: F1 score for private class, U-BA(%): unweighted binary accuracy, U-F1: unweighted F1 score. The cluster size significantly impacts the performance of TM, whereas ITM is more stable having similar performance across different cluster configurations.

| Cluster size | Model | Performance | F1-public | F1-private | U-BA | U-F1 |
|---|---|---|---|---|---|---|
| 10 | Priv×TM | avg | .908(.003) | .736 (.008) | 86.39 (.412) | .822 (.005) |
| | Priv×TM | mdn | .908 | .737 | 86.33 | .822 |
| | Priv×ITM | avg | .908 (.003) | .740 (.007) | 86.42 (.394) | .824 (.005) |
| | Priv×ITM | mdn | .908 | .739 | 86.36 | .823 |
| 20 | Priv×TM | avg | .898 (.004) | .702 (.012) | 84.76 (.583) | .800 (.008) |
| | Priv×TM | mdn | .898 | .700 | 84.83 | .799 |
| | Priv×ITM | avg | .909 (.004) | .726 (.016) | 86.37 (.687) | .817 (.010) |
| | Priv×ITM | mdn | .910 | .733 | 86.64 | .822 |
| 30 | Priv×TM | avg | .878 (.025) | .623 (.172) | 81.73 (4.64) | .750 (.098) |
| | Priv×TM | mdn | .885 | .677 | 83.05 | .780 |
| | Priv×ITM | avg | .907(.007) | .718 (.027) | 86.02 (1.145) | .813 (.017) |
| | Priv×ITM | mdn | .907 | .726 | 86.19 | .817 |

**Table 5:** Average (avg) and median (mdn) performance across 30 different image clustering random seeds with a minimum cluster size $c_{min}/t_{min}$ of 30 and 50 on VISPR [8] dataset. We test for bigger cluster sizes as the size of the VISPR dataset used for clustering is twice as large as for the PrivacyAlert case. Key: F1-public: F1 score for public class, F1-private: F1 score for private class, U-BA(%): unweighted binary accuracy, U-F1: unweighted F1 score. Priv×TM is more susceptible to the choice of cluster size, whereas Priv×ITM shows a smaller drop in performance with increasing cluster size.

| Cluster size | Model | Performance | F1-public | F1-private | U-BA | U-F1 |
|---|---|---|---|---|---|---|
| 30 | Priv×TM | avg | .822(.093) | .883 (.032) | 86.00 (4.916) | .853 (.062) |
| | Priv×TM | mdn | .846 | .891 | 87.23 | .868 |
| | Priv×ITM | avg | .841 (.006) | .888 (.004) | 86.88 (.457) | .865 (.005) |
| | Priv×ITM | mdn | .840 | .889 | 86.84 | .864 |
| 60 | Priv×TM | avg | .546 (.087) | .777 (.020) | 70.25 (3.537) | .662 (.052) |
| | Priv×TM | mdn | .543 | .774 | 69.68 | .658 |
| | Priv×ITM | avg | .766 (.047) | .839 (.032) | 80.97 (3.818) | .803 (.039) |
| | Priv×ITM | mdn | .769 | .843 | 81.30 | .806 |

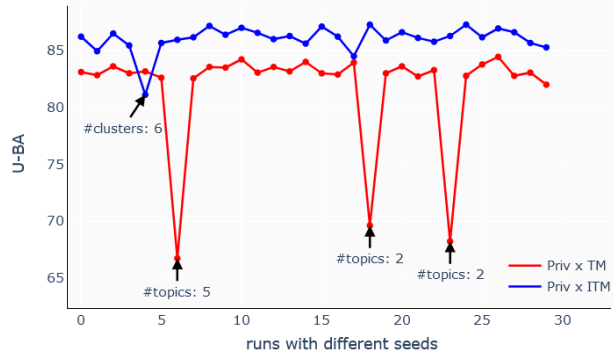

**Fig. 3:** Visualization of accuracy on multiple runs with different random seeds for ITM and TM for $c_{min} = 30$ on PrivacyAlert dataset. Key: U-BA - unweighted binary accuracy We observe that TM, which uses text-based clustering only, is sensitive to random seed selection. The seed choice causes TM to discover a small number of topics (i.e. 2 or 5 topics vs. an average of 20.7(26.26) topics(clusters) for TM(ITM) respectively), which leads to a U-BA lower than 70%.

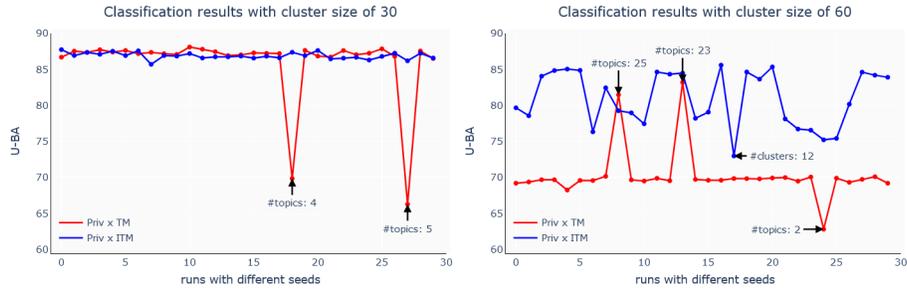

**Fig. 4:** Visualization of accuracy on multiple runs with different random seeds for ITM and TM for $c_{min}/t_{min}$ of 30 and 60 on the VISPR [8] dataset. Key: U-BA - unweighted binary accuracy We observe that TM, which uses text-based clustering only, is sensitive to seed selection. For $t_{min} = 60$, TM frequently discovers 4 topics (which leads to a U-BA lower than 70%) and rarely more than 20 topics, whereas ITM finds on average 22 clusters.

## I    Cluster size vs. Privacy

We analyze the impact of different minimum cluster sizes $(c_{min}, t_{min})$ on the distribution of private images for the PrivacyAlert [12] dataset in both image-based clusters and text-based clusters (i.e. topics) obtained by applying the clustering algorithm (HDBSCAN) on the images and the tag-based representation of images, respectively. We report the distribution of privacy levels of clusters and

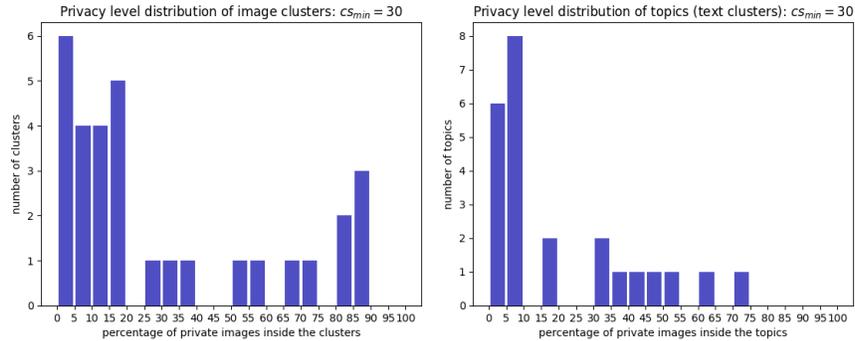

**Fig. 5:** Identified clusters and their privacy level with a minimum cluster size of 30 on the PrivacyAlert [12] dataset.

topics in Fig. 5 and Fig. 6.

**Setup** $c_{min}, t_{min} = 30$: For this configuration, we identify six private image clusters and only one private topic (with a privacy level higher than 70%). Moreover, we observe that directly performing text-based clustering results in more clusters with a high presence of both public and private images than for image clusters. Using the text modality also leads to more images being categorized as outliers ($\simeq$49% of data) than for the image modality ($\simeq$31% of data). This can be due to the fact that text descriptions of images sometimes are generic or they might not include sufficient details that can help better differentiate between samples. Image modality provides richer information (visual cues such as color, texture, shape) which makes it easier to identify underlying patterns and find more specialized clusters.

**Setup** $c_{min}, t_{min} = 10$: The number of the identified clusters increases for both image and text modality but the size of clusters in general is much smaller. For example, for the image clusters we have 3 out of the 11 private clusters with at most 15 samples labeled as private whereas for $c_{min} = 30$ the smallest private cluster has 47 images labeled as private. Although the number of private clusters is different between the setup with $c_{min} = 30$ and $c_{min} = 10$, the corresponding private clusters cover a similar number of private images: 451 for $c_{min} = 30$ and 460 for $c_{min} = 10$. In general, the clusters that have been identified for $c_{min} = 30$ are further split into sub-clusters for $c_{min} = 10$ where images are extremely similar (i.e. same body pose, almost the same outfits, similar camera angle). For the image modality, we also observe that $c_{min} = 10$ produces more clusters with privacy levels in the range 40%-60% than for $c_{min} = 30$. Furthermore, the image modality identifies more private clusters and less uncertain clusters than the text modality. Also in this configuration, the text modality leads to more outliers ($\simeq$43% of data) than for the image modality ($\simeq$32% of data).

The reduced size of clusters makes the validity of the cluster type (i.e. private or not) less reliable than the bigger clusters as it is based on a smaller sample of

individuals' opinions. It is important to note that our objective is to comprehend what type of image content people perceive as private rather than just focusing on the content in the dataset. Thus, reducing the uncertainty of cluster type should also be considered as it allows us to form a clearer understanding of what is private. One way to achieve this is to increase the minimum cluster size in order to capture the opinion of a wider sample of the population. With a bigger cluster size, we observed less uncertain clusters and that clusters contain images that represent similar scenes but with diverse visuals, favoring generalization to new data.

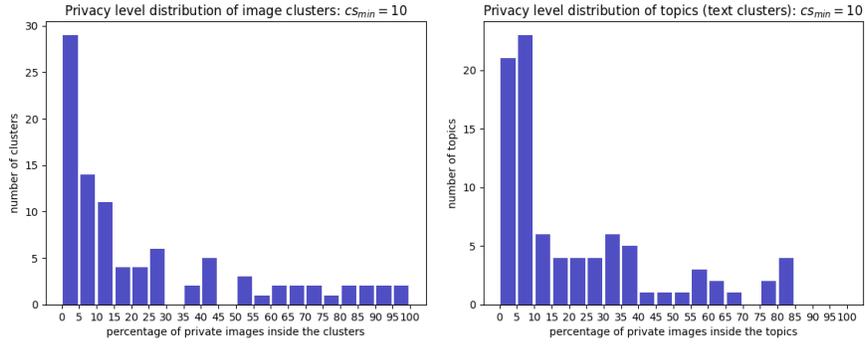

**Fig. 6:** Identified clusters and their privacy level with a minimum cluster size of 10 on the PrivacyAlert dataset.

## J    Example of predictions made by Priv×ITM

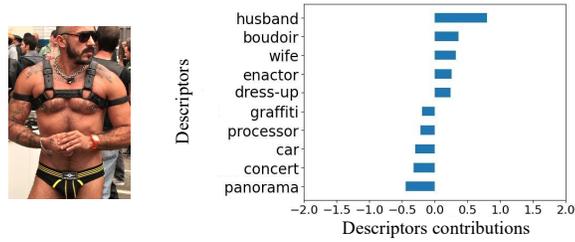

**(a)** GT label: private; Prediction: private

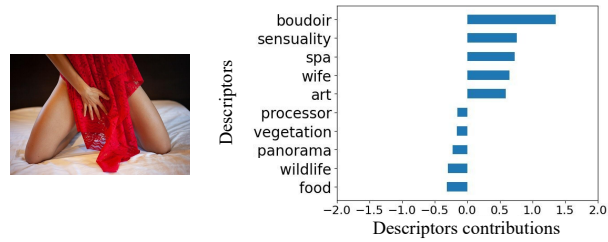

**(b)** GT label: private; Prediction: private

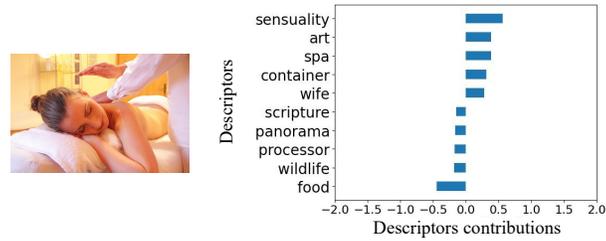

**(c)** GT label: private; Prediction: private

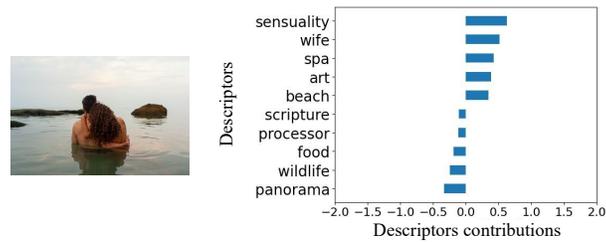

**(d)** GT label: private; Prediction: private

**Fig. 7:** Examples of predictions made by the Priv×ITM on PrivacyAlert [12] dataset. We visualize the top-5 positive and negative descriptors' contributions for each decision. Positive/negative contributions indicate the presence/absence in the image of the content represented by descriptors.

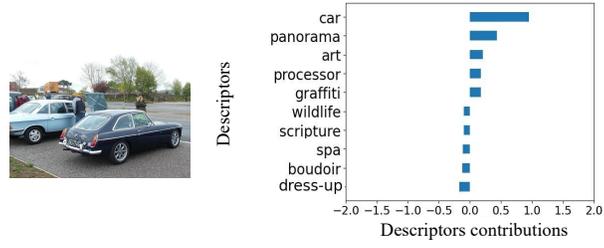

**(a)** GT label: public; Prediction: public

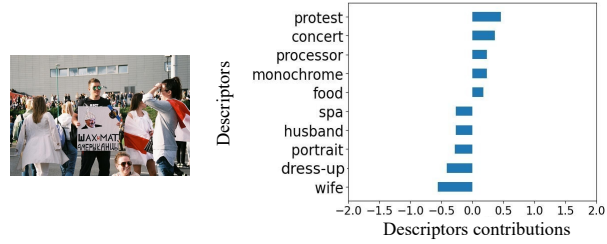

**(b)** GT label: public; Prediction: public

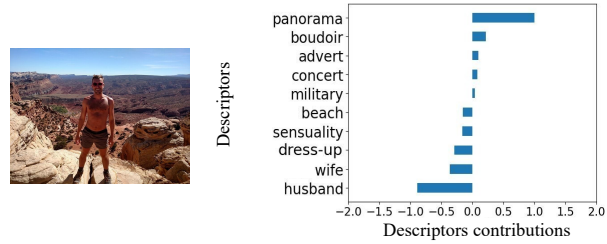

**(c)** GT label: private; Prediction: public

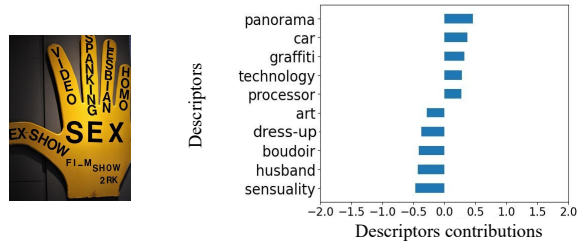

**(d)** GT label: private; Prediction: public

**Fig. 8:** Examples of predictions made by the Priv×ITM on PrivacyAlert [12] dataset. We visualize the top-5 positive and negative descriptors' contributions for each decision. Positive/negative contributions indicate the presence/absence in the image of the content represented by descriptors.

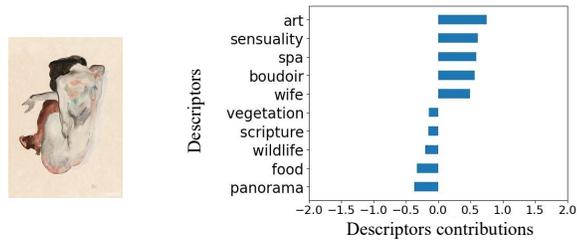

(a) GT label: public; Prediction: private

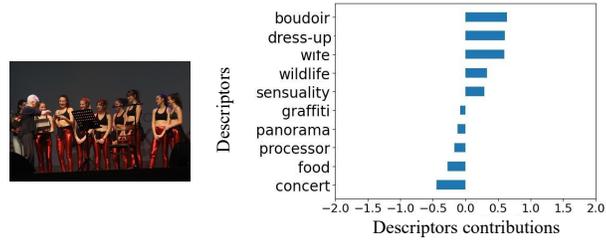

(b) GT label: public; Prediction: private

**Fig. 9:** Examples of predictions made by the Priv×ITM on PrivacyAlert [12] dataset. We visualize the top-5 positive and negative descriptors' contributions for each decision. Positive/negative contributions indicate the presence/absence in the image of the content represented by descriptors.

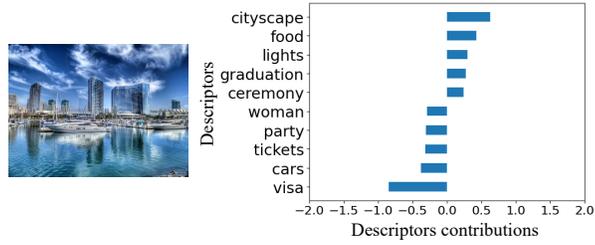

**(a)** GT label: public; Prediction: public

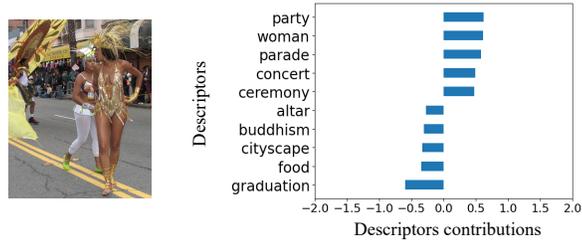

**(b)** GT label: private; Prediction: private

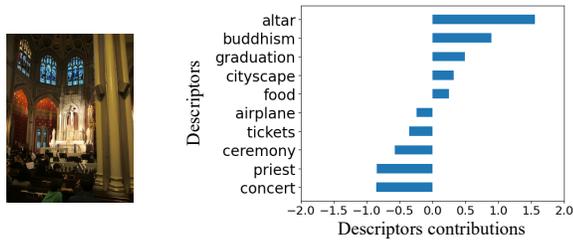

**(c)** GT label: private; Prediction: public

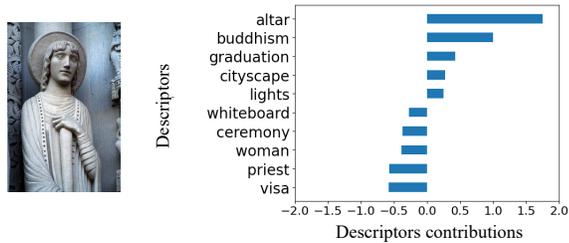

**(d)** GT label: public; Prediction: public

**Fig. 10:** Examples of predictions made by the Priv×ITM on VISPR [8] dataset. We visualize the top-5 positive and negative topics' contributions for each decision. Positive/negative contributions indicate the presence/absence in the image of the content represented by descriptors.

# K    ChatGPT4 concepts generation

We prompt ChatGPT4 with the following instruction to generate a set of concepts that an image privacy classifier should consider in images:

Input: *I want to create a privacy classifier for images, such that I can filter out images to upload to web. What concepts should the classifier look for in private images?*

The answer generated by ChatGPT4 is the following:

Answer: *Creating a privacy classifier for images involves identifying and categorizing content that may be sensitive or personal. Here are some key concepts and features your classifier should look for in images to determine their privacy level:*

1. ***Faces and Personal Identifiers****: The presence of faces can be a strong indicator of private content. Beyond just detecting faces, consider identifying facial expressions, emotions, or specific individuals (if applicable and ethical). This can help determine the sensitivity of the images.*
2. ***License Plates and Vehicle Identification****: Images containing visible license plates or other vehicle identifiers should be flagged, as they can reveal location histories or personal habits.*
3. ***Home Interiors and Exteriors****: Photos that show the inside or outside of homes can inadvertently reveal a person's living situation, personal taste, and location. Recognizing features like house numbers, distinctive architectural details, or even the layout of a home can be crucial.*
4. ***Text and Sensitive Information****: Text detection (OCR) is important for recognizing sensitive information such as names, addresses, phone numbers, email addresses, or any textual content that might be private.*
5. ***Geolocation Data****: If images contain embedded geolocation data (metadata), these can compromise privacy by revealing where the photo was taken, potentially pinpointing someone's home, workplace, or frequently visited locations.*
6. ***Personal Belongings****: Images that show high-value items, personal gadgets, or unique possessions can be sensitive if shared publicly.*
7. ***Children and Minors****: Special care should be taken to identify images of children and minors, as these are generally considered more sensitive, and sharing such images can have legal implications.*
8. ***Health-Related Information****: Photos showing medical reports, medications, or any health-related equipment should be flagged as private due to their sensitive nature.*
9. ***Workplace and Sensitive Locations****: Recognizing images taken in workplaces, especially if they include visible badges, work IDs, or sensitive documents, is important. Additionally, images from locations like schools, banks, or government buildings should be considered sensitive.*

10. **Intimate and Private Moments**: *Detecting images depicting intimate moments or private gatherings can prevent personal moments from being shared unintentionally.*

We train multiple interpretable classifiers, composed of one linear layer whose neurons represent the concepts generated by ChatGPT4. Given that the generated concepts are generic, we explore various strategies:

- Priv×ChatGPT4†: using the 10 concepts highlighted in bold in the ChatGPT4's answer
- Priv×ChatGPT4✁: using the concepts obtained after splitting the names of the concepts generated initially. The splitting is performed manually and 17 concepts are obtained. For example, the concept *license plates and vehicle identification* is split into the *license plates* and *vehicle identification*.
- Priv×ChatGPT4✍: the set of concepts used for Priv×ChatGPT4✁ is further enhanced by manually selecting concrete concepts provided as examples by ChatGPT4. For example, from the initial concept *text and sensitive information* the following *name, address, phone number, email address* are selected. A total of 33 concepts are obtained.
- Priv×ChatGPT4✍+: the set of concepts used for Priv×ChatGPT4✍ is further enhanced by manually adding the concepts for nudity and political preferences using the answer generated with a follow-up prompt to ChatGPT4, for a total of 45 concepts. The prompt and answer are provided below.

Since nudity and political preferences were not covered by the concepts initially generated, the follow-up prompt is given to ChatGPT4:

Input: *What about nudity and political preferences?*
The generated answer is:

Answer: *Nudity and political preferences are indeed crucial categories to consider when designing a privacy classifier for images. Here's how you can handle these sensitive categories:*

1. **Nudity Detection**:
   *Explicit Content: Images containing nudity or explicit content should be flagged as highly sensitive. This includes partial or full nudity, which may not be appropriate for public sharing.*
   *Context Awareness: The classifier should also understand the context in which the nudity appears. For example, art depicting nudity might be treated differently from personal photos. Contextual understanding can help in reducing false positives and managing cultural sensitivities.*
2. **Political Preferences**:
   *Symbols and Signage: Detect political symbols, flags, banners, or clothing that indicate political affiliations or sentiments. This might include logos of political parties, campaign materials, or protest signs.*
   *Contextual Clues: Understanding the context around certain images, like*

*gatherings at a political rally, voting locations, or even private discussions captured in an image, can be indicative of political preferences.*
*Text Analysis: In addition to visual cues, text appearing in images (via signs or documents) can also provide clues about political leanings and should be treated with sensitivity.*

## L  Datasets

**PrivacyAlert.** The authors of PrivacyAlert [12] defined a privacy taxonomy of 10 categories based on previous human-centered studies [5, 8]: *Nudity/Sexual, Other people, Unorganized home, Violence, Medical, Drinking/Party, Appearance/Facial expression, Bad character/Unlawful criminal, Religion/Culture, Personal information.* Relevant images for each category were retrieved using Flickr API search queries. The images were annotated through Amazon Mechanical Turk (AMT) with the following instruction: *Assume you have taken these photos, and you are about to upload them on your favorite social network or content sharing site. Please tell us whether these images are either private or public in nature. Assume that the people in the photos are those that you know.*

**VISPR.** The 68 privacy-related attributes of VISPR [8] dataset were defined using different sources, such as EU Data Protection Directive guidelines, social networks sharing policies, and attributes derived from manual analysis of images shared online. The dataset is split into training (10k images), validation (4.2k images), and testing (8k images). The VISPR authors surveyed 305 users via AMT to assess the privacy preferences for the attributes. Users rated their privacy concerns for online attribute sharing on a 5-point scale (1=no privacy violation, 5=extreme privacy violation). VISPR needs to be enriched with binary labels to be used with our pipeline. Thus, we compute an image privacy risk score $R^I$ for an image $I$ as $R^I = \max(\mathbf{z} \odot \mathbf{u})$, $R^I \in \mathbb{R}^{[0.5,5]}$, where $\mathbf{z} \in \{0,1\}^{|A|}$ is a $k$-hot vector with the ground-truth attribute annotation, $\mathbf{u} \in [0.5,5]^{|A|}$ is the users' average privacy preferences for attributes in $A$, $|A| = 68$, $\max(\cdot)$ yields the maximum element in a vector, and $\odot$ is the element-wise product. A privacy preference of 0.5 is assigned for the *safe* attribute for all users [8]. An image is as privacy-sensitive as its most sensitive attribute. The higher the privacy risk score, the more an image is privacy-sensitive. Then, similar to the approach in VISPR [8], we consider an image $I$ to be *private* if $R^I \geq 3$, and *public* otherwise. Our method does not directly use the attributes annotation, but only indirectly via the generated ground-truth binary labels. We obtain a $\simeq$ 58-42% private-public class distribution for both training and test sets.




\end{document}